\documentclass[10pt,twocolumn,twoside] {IEEEtran}
\usepackage{amsmath,graphicx}
\usepackage[english]{babel}
\usepackage[latin1]{inputenc}
\usepackage{tikz}
\usepackage{url}
\usepackage{xspace}
\usepackage{amsthm}
\usepackage{amssymb}
\usepackage{algorithm}
\usepackage{algorithmicx}
\usepackage{cite}

\newtheorem{lem}{Lemma}

\theoremstyle{definition}
\newtheorem{exmp}{Example}

\renewcommand{\d}{\:{\rm{d}}} 

\newcommand{\Tr}{\operatorname{Tr}}

\newcommand{\ISE}{\operatorname{ISE}}

\renewcommand{\t}[1]{\mathrm{T}#1}
\newcommand{\N}{\mathcal{N}}

\renewcommand{\baselinestretch}{0.9}

\newcommand{\pluseq}{\stackrel{+}{=}}
\usepackage{cite} % Loading the cite package will result in citation numbers being automatically sorted and properly "ranged". i.e., [1], [9], [2], [7], [5], [6] (without using cite.sty) will become: [1], [2], [5]--[7], [9] (using cite.sty)
\usepackage{stfloats}

\usepackage{subcaption}
\usepackage{multirow}
\usepackage{booktabs}

\graphicspath{{./}{figures/}}
\title{Gaussian Mixture Reduction Using Reverse Kullback-Leibler Divergence }

\author{Tohid Ardeshiri, Umut Orguner, Emre \"{O}zkan

\thanks{T. Ardeshiri and E. \"{O}zkan are with the Department of Electrical Engineering, Link\"{o}ping University, 58183 Link\"{o}ping, Sweden, e-mail: tohid,emre@isy.liu.se.}% <-this % stops a space
\thanks{ U. Orguner is  with Department of Electrical and Electronics Engineering, Middle East Technical University, 06800 Ankara, Turkey, email: umut@metu.edu.tr .}% <-this %
}
\begin{document}
\setlength{\interdisplaylinepenalty}{2500}
\maketitle
%%=============================================================================
%%===============================  New SECTION  ===============================
%%=============================================================================
\noindent

\begin{abstract}
We propose a greedy mixture reduction algorithm which is capable of pruning mixture components as well as merging them based on the Kullback-Leibler divergence (KLD).
The algorithm is distinct from the well-known Runnalls' KLD based method since it is not restricted to merging operations.
The capability of pruning (in addition to merging) gives the algorithm the ability of preserving the peaks of the original mixture during the reduction.
Analytical approximations are derived to circumvent the computational intractability of the KLD which results in a computationally efficient method. The proposed algorithm is compared with  Runnalls' and Williams' methods in two numerical examples, using both simulated and real world data. The results indicate that the performance and computational complexity of the proposed approach make it an efficient alternative to existing mixture reduction methods.
%Approximating a Gaussian mixture by another with less components is required in many estimation problems. A greedy solution to this approximation problem is obtained by sequential reduction of the number of components in the mixture density either by deleting one component or via merging two components into one component. In this paper, a greedy mixture reduction algorithm is proposed which selects the optimal choice among the possible choices for  deleting or merging components  via minimizing a cost. The  Kullback-Leibler divergence (KLD) of the mixture with less components to the original mixture is used as the ideal cost in this paper. Since the computation of the exact cost is not practical, analytical approximations of the proposed ideal cost are provided. The proposed algorithm is evaluated in two numerical examples using both simulated data and real-world integrated navigation data.
\end{abstract}

%%=============================================================================
%%===============================  New SECTION  ===============================
%%=============================================================================
 %\vfill
\section{Introduction}
\label{sec:intro}
Mixture densities appear in various problems of the estimation theory. The existing solutions for these problems often require efficient strategies to reduce the number of components in the mixture representation because of computational limits. For example, time series problems can involve an ever increasing number of components in a mixture over time. The Gaussian sum filter \cite{gsum1972} for nonlinear state estimation;
%the multiple model filtering for maneuvering target tracking\cite{blackman1999design}
Multi-Hypotheses Tracking (MHT) \cite{blackman1999design} and Gaussian Mixture Probability Hypothesis Density (GM-PHD) filter \cite{gmphd2006} for multiple target tracking can be listed as examples of the algorithms which require mixture reduction (abbreviated as MR in the sequel) in their implementation.

%harmse2009,  bruneau2010,
Several methods were proposed in the literature addressing the MR problem. %, such as \cite{salmond1989, salmond2009, cu2006,gcu2010, huber2008, tohid2012, Schieferdecker2009, crouse2011}.
In \cite{salmond1989} and \cite{salmond2009}, the components of the mixture were successively merged in pairs for minimizing a cost function.
A Gaussian MR algorithm using homotopy to avoid local minima was suggested in \cite{huber2008}.
Merging statistics for greedy MR for multiple target tracking was discussed in \cite{tohid2012}.
A Gaussian MR algorithm using clustering techniques was proposed in~\cite{Schieferdecker2009}. In~\cite{crouse2011},
Crouse et al. presented a survey of the Gaussian MR algorithms such as West's algorithm~\cite{West1993b}, constraint optimized weight adaptation~\cite{cowa2010}, Runnalls' algorithm~\cite{KLrunnalls2007} and Gaussian mixture reduction via clustering~\cite{Schieferdecker2009} and compared them in detail.

Williams and Maybeck \cite{maybeck2006} proposed using the Integral Square Error (ISE) approach for MR in the multiple hypothesis tracking context. One distinctive feature of the method is the availability of exact analytical expressions for evaluating the cost function between two Gaussian mixtures. Whereas, Runnalls proposed using an upper bound on the Kullback-Leibler divergence (KLD) as a distance measure between the original mixture density and its reduced form at each step of the reduction in \cite{KLrunnalls2007}. The motivation for the choice of an upper bound in Runnalls' algorithm is based on the premise that the KLD between two Gaussian mixtures can not be calculated analytically. Runnalls' approach is dedicated to minimize the KLD from the original mixture to the approximate one, which we refer to here as the \textit{Forward-KLD} (FKLD). This choice of the cost function results in an algorithm, which reduces the number of components only by merging them at each reduction step.

In this paper, we propose a KLD based MR algorithm. Our aim is to find an efficient method for minimizing the KLD from the approximate mixture to the original one, which we refer to as the \textit{Reverse-KLD} (RKLD). The resulting algorithm has the ability to choose between pruning or merging components at each step of the reduction unlike Runnalls' algorithm. This enables, for example, the possibility to prune low-weight components while keeping the heavy-weight components unaltered. Furthermore, we present approximations which are required to overcome the analytical intractability of the RKLD between Gaussian mixtures making the implementation fast and efficient.

The rest of this paper is organized as follows. In Section~\ref{sec:problemdefinition} we present the necessary background required for the MR problem we intend to solve in this paper. Two of the most relevant works and their strengths and weaknesses are described in Section~\ref{sec:relatedwork}.  The proposed solution and its strengths are presented in Section~\ref{sec:RKLD}. Approximations for the fast computation of the proposed divergence are given in Section~\ref{sec:ARKLD}. The proposed MR algorithm using the approximations is evaluated and compared to the alternatives on two numerical examples in Section~\ref{sec:numsim}. The paper is concluded in Section~\ref{sec:conclusion}.

%%=============================================================================
%%===============================  New SECTION  ===============================
%%=============================================================================

\section{Background}
\label{sec:problemdefinition}
A \emph{mixture density} is a convex combination of (more basic) probability densities, see e.g. \cite{bishop2006}. A normalized mixture with $N$ components is defined as
\begin{align}
\label{eq:originalmixture}
p(x)=\sum_{I=1}^{N}{w_{I}q(x;\eta_I)},
\end{align}
where the terms $w_I$ are positive weights summing up to unity, and $\eta_I$ are the parameters of the component density $q(x;\eta_I)$.

The \emph{mixture reduction problem} (MRP) is to find an approximation of the original mixture using fewer components. Ideally, the MRP is formulated as a nonlinear optimization problem where a cost function %
%\footnote{A divergence is a function which establishes the distance of one probability distribution to the other on a statistical manifold \cite{?}.  A divergence measure is a weaker form of a metric, in particular the divergence need not be symmetric and need not satisfy the triangle inequality.}
 measuring the distance between the original and the approximate mixture is minimized. The optimization problem is solved by numerical solvers when the problem is not analytically tractable. The numerical optimization based approaches can be computationally quite expensive, in particular for high dimensional data, and they generally suffer from the problem of local optima~\cite{crouse2011, huber2008,maybeck2006}. Hence, a common alternative solution has been the greedy iterative approach.

In the \emph{greedy} approach, the number of components in the mixture is reduced one at a time. By applying the same procedure over and over again, a desired number of components is reached. To reduce the number of components by one, a \emph{decision} has to be made among two types of operations; namely, the pruning and the merging operations. Each of these operations are considered to be a \emph{hypothesis} in a greedy MRP and is denoted as  $\mathcal{H}_{IJ}$, $I=0,\ldots,N$, $J=1,\ldots,N$, $I\neq J$.

\subsubsection{Pruning}
Pruning is the simplest operation for reducing the number of components in a mixture density. It is denoted with the hypothesis $\mathcal{H}_{0J}$, $J=1,\ldots,N$ in the sequel. In pruning, one component of the mixture is removed and the weights of the remaining components are rescaled such that the mixture integrates to unity. For example, choosing the hypothesis $\mathcal{H}_{0J}$, i.e., pruning component $J$ from~\eqref{eq:originalmixture}, results in the reduced mixture %$(1-w^J)^{-1}\sum_{I=1,I\neq J}^{N}{w^{I}q(x;\eta^I)}$.
\begin{equation}
\label{eq:prunedmixture}
\hat{p}(x|H_{0J})\triangleq \frac{1}{(1-w_J)}\sum_{\substack{I=1\\ I\neq J}}^{N}{w_{I}q(x;\eta_I)}.
\end{equation}
\subsubsection{Merging}
The merging operation approximates a pair of components in a mixture density with a single component of the same type. It is denoted with the hypothesis $\mathcal{H}_{IJ}$, $1\le I\neq J\le N$ in the sequel. In general, an optimization problem minimizing a divergence between the normalized pair of the mixture and the single component is used for this purpose. Choosing the FKLD as the the cost function for merging two components, leads to a moment matching operation.
More specifically, if the hypothesis $\mathcal{H}_{IJ}$ is selected, i.e., if the components $I$ and $J$ are chosen to be merged, the parameters of the  merged component are found by minimizing the divergence from the kernel $w_Iq(x;\eta_I)+w_Jq(x;\eta_J)$ to a single weighted component $(w_I+w_J)q(x;\eta_{IJ})$ as follows:
\begin{equation*}
\eta_{IJ}=\arg \min_{\eta} {D_{KL}\left({\widehat{w}_Iq(x;\eta_I)+\widehat{w}_Jq(x;\eta_J)}\left|\right|q(x;\eta)\right)},
\end{equation*}
where, $\widehat{w}_I=w_I/(w_I+w_J)$, $\widehat{w}_J=w_J/(w_I+w_J)$ and $D_{KL}(p||q)$ denotes the KLD from $p(\cdot)$ to $q(\cdot)$ which is defined as
\begin{equation}
\label{eq:KLDef}
D_{KL}(p||q)\triangleq \int p(x)\log\frac{p(x)}{q(x)} \d x.
\end{equation}
The minimization of the above cost function usually results in the single component $q(x;\eta_{IJ})$ whose several moments are matched to those of the two component mixture. The reduced mixture after merging is then given as
\begin{align}
\hat{p}(x|\mathcal{H}_{IJ})\triangleq \sum_{\substack{K=1\\ K\neq I\\K\neq J}}^{N}{w_{K}q(x;\eta_K)}+(w_I+w_J)q(x;\eta_{IJ}).
\end{align}

There are two different types of greedy approaches in the literature,  namely, local and global approaches.
The local approaches consider only the merging hypotheses $\mathcal{H}_{IJ}$, $1\le I\neq J\le N$. In general, the merging hypothesis $\mathcal{H}_{IJ}$ which provides the smallest divergence $D_{\text{local}}(q(\cdot;\eta_I)||q(\cdot;\eta_J))$ is selected. This divergence considers only the components to be merged and neglects the others. Therefore these methods are called local.  Well-known examples of local approaches are given in~\cite{SalmondSPIE,giw2012}.

In the global approaches, both pruning and merging operations are considered. The divergence between the original and the reduced mixtures, i.e., $D_{\text{global}}(p(\cdot)||\hat{p}(\cdot|\mathcal{H}_{IJ}))$ is minimized in the decision. Because the decision criterion for a global approach involves all of the components of the original mixture, global approaches are in general computationally more costly. On the other hand, since the global properties of the original mixture are taken into account, they provide better performance. In the following, we propose a global greedy MR method that can be implemented efficiently.

%%=============================================================================
%%===============================  New SECTION  ===============================
%%=============================================================================
\section{Related work}
\label{sec:relatedwork}
In this section, we give an overview and a discussion for two well-known global MR algorithms related to the current work.
%%=============================================================================
%%===============================  New SECTION  ===============================
%%=============================================================================
\subsection{Runnalls' Method}
\label{sec:AKL}
Runnalls' method \cite{KLrunnalls2007} is a global greedy MR algorithm that minimizes the FKLD (i.e., $D_{KL}(p(\cdot)||\hat{p}(\cdot)$). Unfortunately, the KLD between two Gaussian mixtures can not be calculated analytically. Runnalls uses an analytical upper-bound for the KLD which can only be used for comparing merging hypotheses. The upper bound $B(I,J)$ for $D_{KL}(p(\cdot)||\hat{p}(\cdot|\mathcal{H}_{IJ}))$, which is given as
\begin{equation}
\label{eq:weightedsum}
\begin{split}
B(I,J)\triangleq &  w_{I}D_{KL}(q(x;\eta_I)||q(x;\eta_{IJ}))\\
&\hspace{1cm}+w_{J}D_{KL}(q(x;\eta_J)||q(x;\eta_{IJ})),
\end{split}
\end{equation}
is used as the cost of merging the components $I$ and $J$
where $q(\cdot,\eta_{IJ})$ is the merged component density. Hence, the original global decision statistics $D_{KL}(p(\cdot)||\hat{p}(\cdot|\mathcal{H}_{IJ}))$ for merging is replaced
with its local approximation $B(I,J)$ to obtain the decision rule as follows:
\begin{align}
\label{eq:mapbij}
i^*,j^*=\arg \min_{1\le I\neq J\le N} B(I,J).
\end{align}
%%=============================================================================
%%===============================  New SECTION  ===============================
%%=============================================================================

\subsection{Williams' Method}
\label{sec:ISE}
Williams and Maybeck proposed a global greedy MRA in~\cite{maybeck2006} where ISE is used as the cost function. ISE between two probability distributions $p(\cdot)$ and $q(\cdot)$ is defined by
\begin{equation}
D_{\ISE}(p||q)\triangleq\int{|p(x)-q(x)|^2\d x}.
\end{equation}
ISE has all properties of a metric such as symmetry and triangle inequality and is analytically tractable for Gaussian mixtures.
Williams' method minimizes $D_{\ISE} (p(\cdot)||\hat{p}(\cdot|\mathcal{H}_{IJ}))$ over all pruning and merging hypotheses, i.e.,
\begin{align}
\label{eq:mapISEij}
i^*,j^*=\arg \min_{\substack{0\le I\le N\\1\le J\le N\\I\neq J}} D_{\ISE}(p(\cdot)||\hat{p}(\cdot|\mathcal{H}_{IJ})).
\end{align}

%%=============================================================================
\subsection{Discussion}
\label{sec:discussion}
In this subsection, first we illustrate the behavior of Runnalls' and Willams' methods on a very simple MR example. Second, we provide a brief discussion on the characteristics observed in the examples along with their implications.
%%=============================================================================
\vspace{0.5cm}
\begin{exmp}
\label{ex:KLDproblem}
Consider the Gaussian mixture with two components given below.
\begin{align}
\label{eq:2Nmixture}
p(x)=w_1\N(x;-\mu,1)+w_2\N(x;\mu,1)
\end{align}
where $\mu\gg 2$ and $w_1+w_2=1$. We would like to reduce the mixture to a single component and hence we consider the two pruning hypotheses $\mathcal{H}_{01}$, $\mathcal{H}_{02}$ and the merging hypothesis $\mathcal{H}_{12}$. The reduced mixtures under these hypotheses are given as
\begin{subequations}
\label{eq:hypexampleisekl}
\begin{align}
\hat{p}(x|\mathcal{H}_{01})=&\N(x;\mu,1),\\
\hat{p}(x|\mathcal{H}_{02})=&\N(x;-\mu,1),\label{eq:hypexampleisekl-H02}\\
\hat{p}(x|\mathcal{H}_{12})=&\N(x;\overline{\mu},\Sigma), \label{eq:Ex1merging}
\end{align}
\end{subequations}
where $\overline{\mu}$ and $\Sigma$ are computed via moment matching as
\begin{subequations}
\begin{align}
\overline{\mu}=&(w_2-w_1)\mu,\\
\Sigma=& 1+4w_1w_2\mu^2.
\end{align}
\end{subequations}
Noting that the optimization problem
\begin{align}
\mu^*, P^*=\arg\min_{\mu,P} D_{KL}(p(\cdot)||\N(\cdot,\mu,P))
\end{align}
has the solution $\mu^*=\overline{\mu}$ and $P^*=\Sigma$, we see that the density $\hat{p}(x|\mathcal{H}_{12})$ is the best reduced mixture with respect to FKLD. Similarly, Runnalls' method would select $\mathcal{H}_{12}$ as the best mixture reduction hypothesis because it considers only the merging hypotheses.\hfill $\blacksquare$
\end{exmp}
\begin{exmp}
\label{ex:ISEproblem} We consider the same MR problem in Example~\ref{ex:KLDproblem} with the Williams' method. The ISE between the original mixture and the reduced mixture can be written as
\begin{align}
D_{\ISE}(p(\cdot)||\hat{p}(\cdot|\mathcal{H}_{IJ}))\pluseq&\int{\hat{p}(x|\mathcal{H}_{IJ})^2\d x}\nonumber\\
&-2\int{p(x)\hat{p}(x|\mathcal{H}_{IJ})\d x} \label{eq:ISEexample}
\end{align}
where the sign $\pluseq$ means equality up to an additive constant independent of $\mathcal{H}_{IJ}$. Letting   $\hat{p}(x|\mathcal{H}_{IJ})=\N(x;\mu_{IJ},\Sigma_{IJ})$, we can calculate $D_{\ISE}(p(\cdot)||\hat{p}(\cdot|\mathcal{H}_{IJ}))$ using~\eqref{eq:ISEexample} as
\begin{align}
D_{\ISE}(p(\cdot)||\hat{p}(\cdot|\mathcal{H}_{IJ}))\pluseq&\N(\mu_{IJ};\mu_{IJ},2\Sigma_{IJ})\nonumber\\
&-2w_1\N(\mu_{IJ};-\mu,1+\Sigma_{IJ})\nonumber\\
&-2w_2\N(\mu_{IJ};\mu,1+\Sigma_{IJ}).
\end{align}
Hence, for the hypotheses listed in~\eqref{eq:hypexampleisekl}, we have
\begin{align}
D_{\ISE}(p(\cdot)||\hat{p}(\cdot|\mathcal{H}_{01}))\pluseq&(1-2w_2)\N(\mu;\mu,2)\nonumber\\
&-2w_1\N(\mu;-\mu,2)\\
=&(1-2w_2)\N(0;0,2)\nonumber\\
&-2w_1\N(\mu;-\mu,2)\\
D_{\ISE}(p(\cdot)||\hat{p}(\cdot|\mathcal{H}_{02}))\pluseq&(1-2w_1)\N(-\mu;-\mu,2)\nonumber\\
&-2w_2\N(-\mu;\mu,2)\\
=&(1-2w_1)\N(0;0,2)\nonumber\\
&-2w_2\N(\mu;-\mu,2)\\
D_{\ISE}(p(\cdot)||\hat{p}(\cdot|\mathcal{H}_{12}))\pluseq&\N(\overline{\mu};\overline{\mu},2\Sigma)\nonumber\\
&-2w_1\N(\overline{\mu};-\mu,1+\Sigma)\nonumber\\
&-2w_2\N(\overline{\mu};\mu,1+\Sigma)\\
=&\N(0;0,2\Sigma)\nonumber\\
&-2w_1\N(2w_2\mu;0,1+\Sigma)\nonumber\\
&-2w_2\N(2w_1\mu;0,1+\Sigma)
\end{align}
Restricting ourselves to the case where $w_1=w_2=1/2$ and $\mu$ is very large, we can see that
\begin{align}
D_{\ISE}(p(\cdot)||\hat{p}(\cdot|\mathcal{H}_{01}))\approx& -\frac{1}{\sqrt{4\pi}}e^{-\mu^2}\\
D_{\ISE}(p(\cdot)||\hat{p}(\cdot|\mathcal{H}_{02}))\approx&-\frac{1}{\sqrt{4\pi}}e^{-\mu^2}\\
D_{\ISE}(p(\cdot)||\hat{p}(\cdot|\mathcal{H}_{12}))\approx&\frac{1}{\sqrt{2\pi}\mu}\left(\frac{1}{\sqrt{2}}-\frac{2}{\sqrt{e}}\right)\approx-\frac{0.51}{\sqrt{2\pi}\mu}.
\end{align}
For sufficiently large $\mu$ values, it is now easily seen that
\begin{align}
D_{\ISE}(p(\cdot)||\hat{p}(\cdot|\mathcal{H}_{12}))<& D_{\ISE}(p(\cdot)||\hat{p}(\cdot|\mathcal{H}_{01}))\nonumber\\
&=D_{\ISE}(p(\cdot)||\hat{p}(\cdot|\mathcal{H}_{02})).
\end{align}
Hence, under the aforementioned conditions, the merging hypothesis is selected by Williams' method.\hfill $\blacksquare$
\end{exmp}
%%=============================================================================
As basically illustrated in Example~\ref{ex:KLDproblem}, FKLD has a tendency towards the merging operation no matter how separated the components of the original mixture are. Similarly Runnalls' method considers only the merging operations ruling out the pruning hypotheses from the start. The significant preference towards the merging operation tends to produce reduced mixtures which may have significant support over the regions where the original mixture have negligible probability mass. This is called as the \emph{zero-avoiding} behavior of the KLD in the literature~\cite[p. 470]{bishop2006}. Such a tendency may be preferable in some applications such as minimum mean square error (MMSE) based estimation. On the other hand, it may also lead to a loss of the important details of the original mixture, e.g., the mode, which might be less desirable in applications such as maximum a posteriori (MAP) estimation. In such applications, having the pruning operation as a preferable alternative might preserve significant information while at the same time keeping a reasonable level of overlap between the supports of the original and the reduced mixtures.

Example~\ref{ex:ISEproblem} illustrated that a similar tendency toward merging (when the components are far from each other) can appear in the case of Williams' method for some specific cases (weights being equal). It must be mentioned here that, in Example~\ref{ex:ISEproblem}, if the weights of components were not the same, Williams' method would select to prune the component with the smaller weight. Therefore, the tendency toward merging (when the components are far from each other) is significantly less in Williams' method than in FKLD and Runnalls' method. It is also important to mention that, in some target tracking algorithms such as MHT and GM-PHD filters, mixtures with some components with identical weights are commonly encountered.

Williams' method, being a global greedy approach to MR, is computationally quite expensive for mixture densities with many components. The computational burden results from the following facts. Reducing a mixture with $N$ components to a mixture with $N-1$ components involves $\mathcal{O}(N^2)$ hypotheses. Since computational load of calculating the ISE between mixtures of $N$ and $N-1$ components is $\mathcal{O}(N^2)$, reducing a mixture with $N$ components to a mixture with $N-1$ components has the computation complexity $\mathcal{O}(N^4)$ with Williams' method. On the other hand, using the upper bound~\eqref{eq:weightedsum}, Runnalls' method avoids the computations associated with the components which are not directly involved in the merging operation resulting in just $\mathcal{O}(N^2)$  computations for the same reduction. Another disadvantage of Williams' method is that the ISE does not scale up with the dimension nicely, as pointed out in an example in~\cite{KLrunnalls2007}.

%%=============================================================================
%%===============================  New SECTION  ===============================
%%=============================================================================
\section{Proposed Method}
\label{sec:RKLD}

We here propose a greedy global MR algorithm based on KLD which can give credit to pruning operations and avoid merging (unlike Runnalls' method) when components of the mixture are far away from each other. The MR method we propose minimizes the KLD from the reduced mixture to the original mixture, i.e., RKLD. Hence we solve the following optimization problem.
\begin{align}
\label{eq:RKLD}
I^*,J^*=\arg \min_{\substack{0\le I\le N\\1\le J\le N\\I\neq J}} D_{KL}(\hat{p}(x|\mathcal{H}_{IJ})||p(x)).
\end{align}
By using the RKLD as the cost function, we aim to enable pruning and avoid the ever merging behavior of Runnalls' method unless it is found necessary. We illustrate the characteristics of this cost function in MR with the following examples.

\begin{exmp}
\label{ex:mergingvspruning1}
We consider the same MR problem
in Example~\ref{ex:KLDproblem} and~\ref{ex:ISEproblem} when $\mu>0$ is very small. When $\mu$ is sufficiently close to zero, we can express $D_{KL}(\hat{p}(x|\mathcal{H}_{IJ})||p(x))$ using a second-order Taylor series approximation around $\mu=0$ as follows:
\begin{align}
D_{KL}(\hat{p}(x|\mathcal{H}_{IJ})||p(x))\approx \frac{1}{2}c_{IJ}\mu^2
\end{align}
where
\begin{align}
c_{IJ}\triangleq \left.\frac{\partial^2}{\partial \mu^2}D_{KL}(\hat{p}(x|\mathcal{H}_{IJ})||p(x)) \right|_{\mu=0}.
\end{align}
This is because, when $\mu=0$, both $p(\cdot)$ and $\hat{p}(\cdot|\mathcal{H}_{IJ})$ are equal to $\N(\cdot;0,1)$
and therefore $D_{KL}(\hat{p}(x|\mathcal{H}_{IJ})||p(x))=0$. Since $D_{KL}(\hat{p}(x|\mathcal{H}_{IJ})||p(x))$ is minimized at $\mu=0$, the first derivative of $D_{KL}(\hat{p}(x|\mathcal{H}_{IJ})||p(x))$ also vanishes at $\mu=0$.
The second derivative term $c_{IJ}$ is given by tedious but straightforward calculations as
\begin{align}
c_{IJ}=\int \frac{\left(\left.\frac{\partial}{\partial\mu}p(x)\right|_{\mu=0}-\left.\frac{\partial}{\partial\mu}\hat{p}(x|\mathcal{H}_{IJ})\right|_{\mu=0}\right)^2}{\N(x;0,1)}\d x.
\end{align}
Using the identity
\begin{align}
\frac{\partial}{\partial\mu}\N(x;\mu,1)=(x-\mu)\N(x;\mu,1)
\end{align}
we can  now  calculate $c_{01}$, $c_{02}$ and $c_{12}$ as
\begin{subequations}
\begin{align}
c_{01}=&4w_1^2\\
c_{02}=&4w_2^2,\\
c_{12}=&0.
\end{align}
\end{subequations}
Hence, when $\mu$ is sufficiently small, the RKLD cost function results in the selection of merging operation similar to FKLD and Runnalls' method.\hfill$\blacksquare$
\end{exmp}
%%=============================================================================
\vspace{0.3cm}
\begin{exmp}
\label{ex:mergingvspruning2}
We consider the same MR problem
in Example~\ref{ex:KLDproblem} and~\ref{ex:ISEproblem} when $\mu$ is very large. In the following, we calculate the RKLDs for the hypotheses given in~\eqref{eq:hypexampleisekl}.
\begin{itemize}
\item $\mathcal{H}_{01}$ and $\mathcal{H}_{02}$: We can write the RKLD for $\mathcal{H}_{01}$ as
\begin{align}
D_{KL}(\hat{p}(\cdot|\mathcal{H}_{01})||p(\cdot))=&-E_{\mathcal{H}_{01}}[\log p(x)]-H(\hat{p}(\cdot|\mathcal{H}_{01})),\label{eq:RKLDH01}
\end{align}
where the notation $E_{\mathcal{H}_{IJ}}[\cdot]$ denotes the expectation operation on the argument with respect to the density function $\hat{p}(\cdot|\mathcal{H}_{IJ})$ and $H(\cdot)$ denotes the entropy of the argument density. Under the assumption that $\mu$ is very large, we can approximate the expectation $E_{\mathcal{H}_{01}}[\log p(x)]$ as
\begin{align}
E_{\mathcal{H}_{01}}&[\log p(x)]\nonumber\\
\triangleq&\int \N(x;\mu,1)\nonumber\\
&\times\log\left(w_1\N(x;-\mu,1)+w_2\N(x;\mu,1)\right)\d x\label{eq:ExactIntegral1}\\
\approx&\int \N(x;\mu,1)\log\left(w_2\N(x;\mu,1)\right)\d x\\
=&\log w_2-\log\sqrt{2\pi}-\frac{1}{2}
\end{align}
where we used the fact that over the effective integration range (around the mean $\mu$), we have
\begin{align}
w_2\N(x;-\mu,1)\approx 0.
\end{align}
Substituting this result and the entropy\footnote{Entropy of a Gaussian density $\N(\cdot,\mu,\sigma^2)$ is equal to $\log \sqrt{2\pi e \sigma^2}$.} of $\hat{p}(\cdot|\mathcal{H}_{01})$ into~\eqref{eq:RKLDH01} along with, we obtain
\begin{align}
D_{KL}(\hat{p}(\cdot|\mathcal{H}_{01})||p(\cdot))\approx&-\log w_2.
\end{align}
Using similar arguments, we can easily obtain
\begin{align}
D_{KL}(\hat{p}(\cdot|\mathcal{H}_{02})||p(\cdot))\approx& -\log w_1.
\end{align}
Noting that the logarithm function is monotonic, we can also see that the approximations given on the right hand sides of the above equations are upper bounds for the corresponding RKLDs.
\item $\mathcal{H}_{12}$: We now calculate a lower bound on $D_{KL}(\hat{p}(x|\mathcal{H}_{12})|| p(x))$ and show that this lower bound is greater than both $-\log w_1$ and $-\log w_2$ when $\mu$ is sufficiently large.
\begin{align}
& D_{KL}(\hat{p}(x|\mathcal{H}_{12})|| p(x))= -E_{\mathcal{H}_{12}}[\log p(x)]-H(\hat{p}(\cdot|\mathcal{H}_{12}))\label{eq:RKLDH12}
\end{align}
  We consider the following facts
  \begin{subequations}
  \label{eq:identitiesforupperbound}
  \begin{align}
      \N(x;-\mu,1)\ge \N(x;\mu,1)\quad&\text{when}\quad x\le0,\\
      \N(x;\mu,1)\ge \N(x;-\mu,1)\quad&\text{when}\quad x\ge0.
  \end{align}
  \end{subequations}
 Using the identities in~\eqref{eq:identitiesforupperbound} we can obtain a bound on the expectation $E_{\mathcal{H}_{12}}[\log p(x)]$ as
\begin{align}
E_{\mathcal{H}_{12}}&[\log p(x)]\nonumber\\
\triangleq& E_{\mathcal{H}_{12}}[\log\left(w_1\N(x;-\mu,1)+w_2\N(x;\mu,1)\right)]\nonumber\\
\le& \int_{x\le 0}{\N(x;\bar{\mu},\Sigma)\log \left(\N(x;-\mu,1)\right)\d x}\nonumber\\
&+\int_{x>0}{\N(x;\bar{\mu},\Sigma)\log \left(\N(x;\mu,1)\right)\d x}\\
=& -\int_{x\le0}\N(x;\bar{\mu},\Sigma)\bigg(\log \sqrt{2\pi}+\frac{(x+\mu)^2}{2}\bigg)\d x\nonumber\\
&-\int_{x>0}\N(x;\bar{\mu},\Sigma)\bigg(\log \sqrt{2\pi}+\frac{(x-\mu)^2}{2}\bigg)\d x\\
\le&-\log\sqrt{2\pi}-\int_{x\le0}\N(x;\bar{\mu},\Sigma)\frac{x^2+2\mu x+\mu^2}{2}\d x\nonumber\\
&-\int_{x>0}\N(x;\bar{\mu},\Sigma)\frac{x^2-2\mu x+\mu^2}{2}\d x\\
=&-\log\sqrt{2\pi}-\frac{\bar{\mu}^2+\Sigma+\mu^2}{2}\nonumber\\
&-\mu \int_{x\le0}x\N(x;\bar{\mu},\Sigma)\d x\nonumber\\
&+\mu \int_{x>0}x\N(x;\bar{\mu},\Sigma)\d x\\
=&-\log\sqrt{2\pi}-\frac{\bar{\mu}^2+\Sigma+\mu^2}{2}\nonumber\\
&-\mu \int_{x\le0}x\N(x;\bar{\mu},\Sigma)\d x\nonumber\\
&+\mu \bigg(\bar{\mu}-\int_{x\le0}x\N(x;\bar{\mu},\Sigma)\bigg)\d x\\
=&-\log\sqrt{2\pi}-\frac{\bar{\mu}^2+\Sigma+\mu^2}{2}+\mu\bar{\mu}\nonumber\\
&-2\mu \int_{x\le0}x\N(x;\bar{\mu},\Sigma)\d x\\
=&-\log\sqrt{2\pi}-\frac{(\mu-\bar{\mu})^2+\Sigma}{2}\nonumber\\
&-2\mu\bar{\mu}\Phi\left(-\frac{\bar{\mu}}{\sqrt{\Sigma}}\right)+2\mu\sqrt{\Sigma}\phi\left(-\frac{\bar{\mu}}{\sqrt{\Sigma}}\right)\label{eqn:upperboundE12logp}
\end{align}
where we have used the result
\begin{align}
\int_{-\infty}^{\bar{x}} x&\N(x,\mu,\Sigma)\d x\nonumber\\
&=\mu\Phi\left(\frac{\bar{x}-\mu}{\sqrt{\Sigma}}\right)-\sqrt{\Sigma} \phi\left(\frac{\bar{x}-\mu}{\sqrt{\Sigma}}\right).
\end{align}
Here, the functions $\phi(\cdot)$ and $\Phi(\cdot)$ denote the probability density function and  cumulative distribution function of a Gaussian random variable with zero mean and unity variance, respectively.
Substituting the upper bound~\eqref{eqn:upperboundE12logp} into~\eqref{eq:RKLDH12}, we obtain
\begin{align}
D_{KL}&(\hat{p}(x|\mathcal{H}_{12})|| p(x))\nonumber\\
\ge& -\frac{1}{2}-\frac{1}{2}\log\Sigma+\frac{(\mu-\bar{\mu})^2+\Sigma}{2}\nonumber\\
&+2\mu\bar{\mu}\Phi\left(-\frac{\bar{\mu}}{\sqrt{\Sigma}}\right)-2\mu\sqrt{\Sigma}\phi\left(-\frac{\bar{\mu}}{\sqrt{\Sigma}}\right)\\
\approx&-\frac{1}{2}-\frac{1}{2}\log \left(4w_1w_2\mu^2\right)+2\mu^2\bigg(w_1+(w_2-w_1)\nonumber\\
&\times\Phi\left(\frac{w_1-w_2}{\sqrt{4w_1w_2}}\right)-\sqrt{4w_1 w_2}\phi\left(\frac{w_1-w_2}{\sqrt{4w_1w_2}}\right)\bigg)\nonumber\\
=&-\frac{1}{2}-\frac{1}{2}\log\left( 4w_1w_2\mu^2\right)+2g(w_1,w_2)\mu^2
 \label{eq:lowerboundRKLDH12}
\end{align}
for sufficiently large $\mu$ values where we used the definitions of $\bar{\mu}$, $\Sigma$ and the fact that when $\mu$ tends to infinity, we have
\begin{align}
-\frac{\bar{\mu}}{\sqrt{\Sigma}}\rightarrow \frac{w_1-w_2}{\sqrt{4w_1w_2}} .
\end{align}
The coefficient $g(w_1,w_2)$ in~\eqref{eq:lowerboundRKLDH12} is defined as
\begin{align}
g(w_1,w_2)\triangleq& w_1+(w_2-w_1)\Phi\left(\frac{w_1-w_2}{\sqrt{4w_1w_2}}\right)\nonumber\\
&-\sqrt{4w_1 w_2}\phi\left(\frac{w_1-w_2}{\sqrt{4w_1w_2}}\right).
\end{align}
\end{itemize}
%% intermediate steps--------
%\begin{align*}
%&(\mu-\bar{\mu})^2+\Sigma-1\\
%&=((1-w_2+w_1)\mu)^2+(1+4w_1w_2\mu^2)-1\\
%&=4w_1^2\mu^2+4w_1w_2\mu^2\\
%&=4w_1(w_1+w_2)\mu^2=4w_1\mu^2
%\end{align*}
%% end of intermediate steps--------
When $\mu$ tends to infinity, the dominating term on the right hand side of~\eqref{eq:lowerboundRKLDH12} becomes the last  term. By generating a simple plot, we can see that the function $g(w_1,1-w_1)$ is positive for $0<w_1<1$, which makes the right hand side of \eqref{eq:lowerboundRKLDH12} go to infinity as $\mu$ tends to infinity. Consequently, the cost of the merging operation exceeds both of the pruning hypotheses for sufficiently large $\mu$ values. Therefore, the component having the minimum weight is pruned when the components are sufficiently separated.\hfill$\blacksquare$
\end{exmp}

%%=============================================================================
As illustrated in Example~\ref{ex:mergingvspruning2}, when the components of the mixture are far away from each other, RKLD refrains from merging them. This property of RKLD is known as \emph{zero-forcing} in the literature~\cite[p. 470]{bishop2006}.

The RKLD between two Gaussian mixtures is analytically intractable except for trivial cases. Therefore, to be able to use RKLD in MR, approximations are necessary just as in the case of FKLD with Runnalls' method. We propose such approximations of RKLD for the pruning and merging operations in the following section.

%%=============================================================================
%%===============================  New SECTION  ===============================
%%=============================================================================

\section{Approximations for RKLD}
\label{sec:ARKLD}
%Ideally we would like to implement the reduction algorithm by solving the global optimization problem numerically. But such a solution is costly. As a remedy we will use a greedy approach to mixture reduction. Since the Kullback-Leibler divergence between two mixture densities is not analytically tractable we have to use approximations to evaluate the cost of different pruning and merging hypotheses.
In sections~\ref{sec:pruningapprox} and~\ref{sec:mergingapprox}, specifically tailored approximations for the cost functions of pruning and merging hypotheses are derived respectively.
Before proceeding further, we would like to introduce a lemma which is used in the derivations.
\begin{lem}
\label{lem:vub}
Let $q_I(x)$, $q_J(x)$, and $q_K(x)$ be three probability distributions and $w_I$ and $w_J$ two positive real numbers. The following inequality holds
\begin{align}
&\int q_K(x)  \log\frac{q_K(x)}{w_Iq_I(x)+w_Jq_J(x)}\d x\leq\nonumber\\
 &-\log\left( {w_I\exp(-D_{KL}(q_K||q_I))+w_J\exp(-D_{KL}(q_K||q_J))} \right)\label{eq:basicintegral}
\end{align}
\begin{proof}
For the proof see Appendix~\ref{sec:proofoflem:vub}.
\end{proof}
\end{lem}

 %--------------------------

%%=============================================================================

\subsection{Approximations for pruning hypotheses}
\label{sec:pruningapprox}
Consider the mixture density  $p(\cdot)$ defined as
\begin{equation}
\label{eq:originalp}
p(x)=\sum_{I=1}^N{w_I q_I(x)},
\end{equation}
where $q_I(x)\triangleq q(x;\eta_I)$.
Suppose we reduce the mixture in \eqref{eq:originalp} by pruning the $I$th component as follows.
\begin{equation}
\hat{p}(x|\mathcal{H}_{0I})=\frac{1}{1-w_I}\sum_{i \in \{1\cdots N\}-\{I\}}{w_i q_i(x)}.
\end{equation}
An upper bound can be obtained using the fact that the logarithm function is monotonically increasing as
\begin{align}
& D_{KL}(\hat{p}(\cdot|\mathcal{H}_{0I})||p(\cdot))=\int{\hat{p}(x|\mathcal{H}_{0I})\log\frac{\hat{p}(x|\mathcal{H}_{0I})}{p(x)} \d x} \nonumber\\
&= \int{\hat{p}(x|\mathcal{H}_{0I})\log\frac{\hat{p}(x|\mathcal{H}_{0I})}{(1-w_I)\hat{p}(x|\mathcal{H}_{0I})+w_Iq_I} \d x} \nonumber\\
&\le -\log(1-w_I)+\int{\hat{p}(x|\mathcal{H}_{0I})\log\frac{\hat{p}(x|\mathcal{H}_{0I})}{\hat{p}(x|\mathcal{H}_{0I})} \d x} \nonumber\\
&=-\log(1-w_I).\label{eq:crudepruningapproximation}
\end{align}
This upper bound is rather crude when the $I$th component density is close to other component densities in the mixture. Therefore we compute a tighter bound on $D_{KL}(\hat{p}(\cdot|\mathcal{H}_{0I})||p(\cdot))$ using the log-sum inequality~\cite{logsumref}.
Before we derive the upper bound, we first define the following unnormalized density
\begin{equation}
\label{eq:r(x)}
r(x)=\sum_{i \in \{1\cdots N\}-\{I,J\}}{w_i q_i(x)}
\end{equation}
where $J\neq I$ and $\int r(x) \d x=1-w_I-w_J$.

We can rewrite the RKLD between $\hat{p}(\cdot|\mathcal{H}_{0I})$ and $p(\cdot)$ as
%\begin{equation}
%D_{KL}(p(\cdot|\mathcal{H}_{0I})||p(\cdot))\triangleq\int{{p(x|\mathcal{H}_{0I})}\log\frac{p(x|\mathcal{H}_{0I})}{p(x)}\d x}.
%\end{equation}
 %as follows
\begin{align}
D_{KL}&(\hat{p}(\cdot|\mathcal{H}_{0I})||p(\cdot))=\int\frac{1}{1-w_I}(r(x)+w_Jq_J(x)) \nonumber\\
&\times\log\frac{\frac{1}{1-w_I}(r(x)+w_Jq_J(x))}{r(x)+w_Iq_I(x)+w_Jq_J(x)}\d x \nonumber\\
=&-\log(1-w_I)+\frac{1}{1-w_I}\int(r(x)+w_Jq_J(x))\nonumber\\
&\times\log\frac{r(x)+w_Jq_J(x)}{r(x)+w_Iq_I(x)+w_Jq_J(x)}\d x.
\end{align}
Using the log-sum inequality we can obtain the following expression.
\begin{align}
&D_{KL}(\hat{p}(\cdot|\mathcal{H}_{0I})||p(\cdot))\nonumber\\
&\leq -\log(1-w_I)+ \frac{1}{1-w_I}\int{r(x)\log\frac{r(x)}{r(x)}\d x}\nonumber\\
&\hspace{0cm}+\frac{1}{1-w_I}\int{w_Jq_J(x)\log\frac{w_Jq_J(x)}{w_Iq_I(x)+w_Jq_J(x)}\d x} \label{eq:logsumpruning}\nonumber\\
&\hspace{0cm}=-\log(1-w_I)+ \frac{w_J}{1-w_I} \log(w_J) \nonumber\\
&\hspace{0cm}+\frac{w_J}{1-w_I} \int{q_J(x)\log\frac{q_J(x)}{w_Iq_I(x)+w_Jq_J(x)}\d x}.
\end{align}
Applying the result of Lemma~\ref{lem:vub} on the integral, we can write
\begin{align}
D_{KL}&(\hat{p}(\cdot|\mathcal{H}_{0I})||p(\cdot))\le-\log(1-w_I)\nonumber\\
& -\frac{w_J}{1-w_I} \log\left( 1+\frac{w_I}{w_J}\exp(- D_{KL}(q_J||q_I)) \right).
%&\le -\log(1-w_I)+ \frac{w_J}{1-w_I} \log(w_J)+\frac{w_J}{1-w_I}\nonumber\\
%&\times\left[ D_{KL}(q_J||q_I)-\log(w_J\exp(D_{KL}(q_J||q_I))+w_I)\right]
\end{align}
Since $J$ is arbitrary, as long as $J\neq I$, we obtain the upper bound given below.
\begin{align}
& D_{KL}(\hat{p}(\cdot|\mathcal{H}_{0I})||p(\cdot))\le \min_{J \in \{1\cdots N\}-\{I\}}\bigg[-\log(1-w_I) \nonumber\\
&-\frac{w_J}{1-w_I} \log\left( 1+\frac{w_I}{w_J}\exp(- D_{KL}(q_J||q_I)) \right)\bigg].
\end{align}
The proposed approximate divergence for pruning component $I$ will be denoted by $R(0,I)$, where $1\le I\le N$ in the rest of this paper. In the following, we illustrate the advantage of the proposed approximation in a numerical example.
\begin{exmp}
\label{ex:RB02}
Consider Example~\ref{ex:KLDproblem} with the mixture~\eqref{eq:2Nmixture} and the hypothesis~\eqref{eq:hypexampleisekl-H02}.
In Figure~\ref{fig:pruninganalysis} the exact divergence $D_{KL}(p(\cdot|\mathcal{H}_{02})||p(\cdot))$, which is computed numerically, its crude approximation given in~\eqref{eq:crudepruningapproximation} and the proposed approximation $R(0,2)$ are shown for different values of $\mu$ with $w_1=0.8$. Both the exact divergence and the upper bound $R(0,2)$ converge to $-\log(1-w_2)$ when the pruned component has small overlapping probability mass with the other component. The bound $R(0,2)$ brings a significant improvement over the crude bound when the amount of overlap between the components increases.  \hfill$\blacksquare$
%Also, the approximation error is rather small when there is overlapping probability mass.
\begin{figure}[th]
	\centering
		\includegraphics[width=0.45\textwidth]{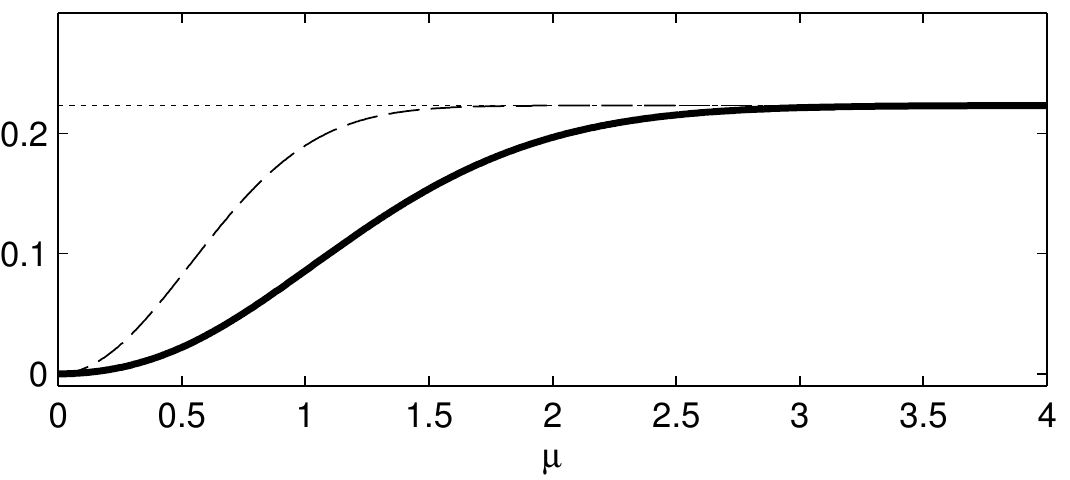}	
		\caption{The exact divergence $D_{KL}(p(\cdot|\mathcal{H}_{02})||p(\cdot))$ (solid black), the crude upper bound $-\log(1-w_2)$ (dotted black) and , the proposed upper bound $R(0,2)$ (dashed black) are plotted for different values of $\mu$.}
\label{fig:pruninganalysis}
\end{figure}
\end{exmp}

%\note{To avoid the costly computations of this bound $\alpha$ can be set to $\frac{w_I}{w_I+w_J}$ which give a looser upper bound given as
%\begin{align}
%& D_{KL}(p(\cdot|\mathcal{H}_{0I})||p(\cdot))\le \min_{J \in \{1\cdots N\}-\{I\}}\left[-\log(1-w_I)\right.\nonumber\\
%&+ \left.\frac{w_J}{1-w_I}\left[ \log\frac{w_J}{w_I+w_I}+\frac{w_I}{w_I+w_J} D_{KL}(q_J||q_I)\right]\right].
%\end{align}
%it requires taking the min of this bound and -log...}

%%=============================================================================
\subsection{Approximations for  merging hypotheses}
\label{sec:mergingapprox}
Consider the problem of merging the $I$th and the $J$th components of the mixture density~\eqref{eq:originalp} where the resulting approximate density is given as follows.
\begin{equation}
\hat{p}(x|\mathcal{H}_{IJ})=w_{IJ}q_{IJ}(x)+\sum_{i \in \{1\cdots N\}-\{I,J\}}{w_i q_i(x)}.
\end{equation}
%Before we derive the upper bound we first define the following function
%\begin{equation}
%r(x)=\sum_{i \in \{1\cdots N\}-\{I,J\}}{w_i q_i(x)}
%\end{equation}
%where $J\neq I\neq K$, $J\neq K$  and $\int r(x) \d x=1-w_I-w_J-w_K$.
We are interested in the RKLD between $\hat{p}(\cdot|\mathcal{H}_{IJ})$ and $p(\cdot)$. Two approximations of this quantity with different accuracy and computational cost are given in sections \ref{sec:vub} and \ref{sec:vapp}.
%\begin{equation}
%D_{KL}(p(\cdot|\mathcal{H}_{IJ})||p(\cdot))\triangleq\int{{p(x|\mathcal{H}_{IJ})}\log\frac{p(x|\mathcal{H}_{IJ})}{p(x)}\d x}.
%\end{equation}
%%=============================================================================
\subsubsection{A simple upper bound}
\label{sec:vub}
%We are interested in the KL divergence between $p(\cdot|\mathcal{H}_{IJ})$ and $p(\cdot)$   given as
%\begin{equation}
%D_{KL}(p(\cdot|\mathcal{H}_{IJ})||p(\cdot))\triangleq\int{{p(x|\mathcal{H}_{IJ})}\log\frac{p(x|\mathcal{H}_{IJ})}{p(x)}\d x}.
%\end{equation}
We can compute a bound on $D_{KL}(\hat{p}(\cdot|\mathcal{H}_{IJ})||p(\cdot))$ as follows:
\begin{align}
D&_{KL}(\hat{p}(\cdot|\mathcal{H}_{IJ})||p(\cdot)) \nonumber\\
=&\int{({r(x)+w_{IJ}q_{IJ}(x)})\log\frac{r(x)+w_{IJ}q_{IJ}(x)}{r(x)+w_Iq_I(x)+w_Jq_J(x)}\d x}\nonumber\\
\leq & \int{r(x)\log\frac{r(x)}{r(x)}\d x}\nonumber\\
&+\int{{w_{IJ}q_{IJ}(x)}\log\frac{w_{IJ}q_{IJ}(x)}{w_Iq_I(x)+w_Jq_J(x)}\d x}\nonumber\\
=& w_{IJ}\log w_{IJ} + w_{IJ}\int{{q_{IJ}(x)}\log\frac{q_{IJ}(x)}{w_Iq_I(x)+w_Jq_J(x)}\d x}\label{eq:integralqIJ}
\end{align}
where the log-sum inequality is used and $r(x)$ is defined in~\eqref{eq:r(x)}. Using Lemma~\ref{lem:vub} for the second term on the right hand side of \eqref{eq:integralqIJ}, we obtain
\begin{align}
&D_{KL}(\hat{p}(\cdot|\mathcal{H}_{IJ})||p(\cdot))\le w_{IJ}\log w_{IJ}-w_{IJ}\times\nonumber\\
&\log\left( {w_I\exp(-D_{KL}(q_{IJ}||q_I))+w_J\exp(-D_{KL}(q_{IJ}||q_J))} \right).\label{eq:revAKLtight}
\end{align}
\subsubsection{A variational approximation}
\label{sec:vapp}
The upper bound derived in the previous subsection is rather loose particularly when the components $q_I$ and $q_J$ are far from each other. This is because of the replacement of the second term in \eqref{eq:integralqIJ} with its approximation in~\eqref{eq:revAKLtight} which is a function of $D_{KL}(q_{IJ}||q_I)$ and $D_{KL}(q_{IJ}||q_J)$.

For a brief description of the problem consider
\begin{align}
\label{eq:DKIJ}
D(q_{IJ}||w_Iq_I+&w_Jq_J)\triangleq \nonumber\\
&\int q_{IJ}(x) \log\frac{q_{IJ}(x)}{w_Iq_I(x)+w_Jq_J(x)}\d x
\end{align}
and its approximation using Lemma~\ref{lem:vub} given as
\begin{align}
-\log\left( {w_I\exp(-D_{KL}(q_{IJ}||q_I))+w_J\exp(-D_{KL}(q_{IJ}||q_J))} \right).\nonumber
\end{align}
The integral involved in the calculation of $D_{KL}(q_{IJ}||q_I)$ grows too much (compared to the original divergence $D(q_{IJ}||w_Iq_I+w_Jq_J)$) over the support of $q_J$ because $q_{IJ}$ shares significant support with $q_J$. Similarly, the integral in $D_{KL}(q_{IJ}||q_J)$ grows too much (compared to the original divergence $D(q_{IJ}||w_Iq_I+w_Jq_J)$)  over the support of $q_I$ because $q_{IJ}$ shares significant support with $q_I$. To fix these problems, we propose the following procedure for approximating $D(q_{IJ}||w_Iq_I+w_Jq_J)$ using the boundedness of the component densities.
In the proposed variational approximation, first, an upper bound on $D(q_{IJ}||w_Iq_I+w_Jq_J)$ is obtained using the log-sum inequality as
\begin{align}
D(&q_{IJ}||w_Iq_I+w_Jq_J)\leq \int{\alpha q_{IJ}\log\frac{\alpha q_{IJ}}{w_Iq_I}\d x}\nonumber\\
 &+\int{(1-\alpha)q_{IJ}\log\frac{(1-\alpha)q_{IJ}}{w_Jq_J}\d x}\\
=&\alpha\int{ q_{IJ}\log\frac{{ q_{IJ}}}{q_I}\d x}+\alpha\log\frac{{\alpha }}{w_I}\nonumber\\
&+  (1-\alpha)\int{ q_{IJ}\log\frac{{ q_{IJ}}}{q_J}\d x} +   (1-\alpha) \log\frac{1-\alpha}{w_J}\label{eq:approxvariationalcost}
\end{align}
where $0<\alpha<1$. Second, a switching function is applied to the integrals in~\eqref{eq:approxvariationalcost} as follows.
\begin{align}
D(q_{IJ}&||w_Iq_I+w_Jq_J) \nonumber\\
\approx& \int{\alpha q_{IJ}(x)\left(1-\frac{q_J(x)}{q_J^{max}}\right)\log\frac{ q_{IJ}(x)}{q_I(x)}\d x}+\alpha\log\frac{{\alpha }}{w_I} \nonumber\\
&+\int{(1-\alpha)q_{IJ}(x)\left(1-\frac{q_I(x)}{q_I^{max}}\right)\log\frac{q_{IJ}(x)}{q_J(x)}\d x}\nonumber\\
&+(1-\alpha) \log\frac{1-\alpha}{w_J}\label{eq:BeforeMinimum}
\end{align}
where $q_I^{max}\triangleq\max_x q_I(x)$. Here, the multipliers $\left(1-\frac{ q_J(x)}{q_J^{max}}\right)$, and $\left(1-\frac{ q_I(x)}{q_I^{max}}\right)$ in the integrands are equal to almost unity everywhere except for the support of $q_{J}$ and $q_I$ respectively, where they shrink to zero. In this way, the excessive growth of the divergences in the upper bound are reduced. Note also that the terms $q_J^{max}$ and $q_I^{max}$ are readily available as the maximum values of the Gaussian component densities. Finally, the minimum of the right hand side of~\eqref{eq:BeforeMinimum} is obtained with respect to $\alpha$. The optimization problem can be solved by taking the derivative with respect to $\alpha$ and equating the result to zero which gives the following optimal value
\begin{equation*}
\alpha^*=\frac{w_I\exp(-V(q_{IJ},q_J,q_I))}{w_I\exp(-V(q_{IJ},q_J,q_I))+w_J\exp(-V(q_{IJ},q_I,q_J)))},
\end{equation*}
where
\begin{align}
V(q_K,q_I,q_J)\triangleq\int{q_{K}(x)\left(1-\frac{q_I(x)}{q_I^{max}}\right)\log\frac{q_{K}(x)}{q_J(x)}\d x}.
\end{align}
A closed form expression for the quantity $V(q_K,q_I,q_J)$  is given in Appendix~\ref{sec:derivationofV}.
Substituting $\alpha^*$ into~\eqref{eq:BeforeMinimum} results in the following approximation for $D(q_{IJ}||w_Iq_I+w_Jq_J)$.
\begin{align}
&D(q_{IJ}||w_Iq_I+w_Jq_J)\approx\nonumber\\
&-\log\left( {w_I\exp(-V(q_{IJ},q_J,q_I))+w_J\exp(-V(q_{IJ},q_I,q_J))}\right)
\end{align}
When we substitute this approximation into~\eqref{eq:integralqIJ}, the proposed approximate divergence for merging becomes
\begin{align}
&D_{KL}(\hat{p}(\cdot|\mathcal{H}_{IJ})||p(\cdot))\approx w_{IJ}\log w_{IJ} -w_{IJ}\times\nonumber\\
&\log\left( {w_I\exp(-V(q_{IJ},q_J,q_I))+w_J\exp(-V(q_{IJ},q_I,q_J))} \right).\label{eq:approxmerging}
\end{align}
The proposed approximate divergence for merging components $I$ and $J$ in~\eqref{eq:approxmerging} will be denoted by $R(I,J)$ in the rest of this paper.
\begin{exmp}
\label{ex:RB12}
Consider Example~\ref{ex:KLDproblem} with the mixture~\eqref{eq:2Nmixture} and the hypothesis~\eqref{eq:Ex1merging}.
In Figure~\ref{fig:merginganalysis}, the exact divergence $D_{KL}(\hat{p}(\cdot|\mathcal{H}_{12})||p(\cdot))$, which is computed numerically, the simple approximation given in~\eqref{eq:revAKLtight} and the proposed approximation $R(1,2)$ are shown for different values of $\mu$ with $w_1=0.8$.

Note that the range at which approximations are interesting to study is where  $D_{KL}(\hat{p}(\cdot|\mathcal{H}_{12})||p(\cdot)) < -\log(1-w_2) $. This is because of the fact that when $D_{KL}(\hat{p}(\cdot|\mathcal{H}_{12})||p(\cdot)) > -\log(1-w_2)$, pruning hypothesis $\mathcal{H}_{02}$ will be selected. The range of axes in Figure~\ref{fig:merginganalysis} is chosen according to the aforementioned range of interest.

The proposed variational approximation $R(1,2)$ performs very well compared to the simple upper bound which grows too fast with increasing $\mu$ values. Furthermore, the points where the curves of exact pruning and merging divergences cross (the crossing point of the solid lines in Figure~\ref{fig:merginganalysis}) is close to the crossing point of the curves of best approximation of these two quantities (the crossing point of the dashed lines in Figure~\ref{fig:merginganalysis}).\hfill$\blacksquare$
\begin{figure}
	\centering
		\includegraphics[width=0.45\textwidth]{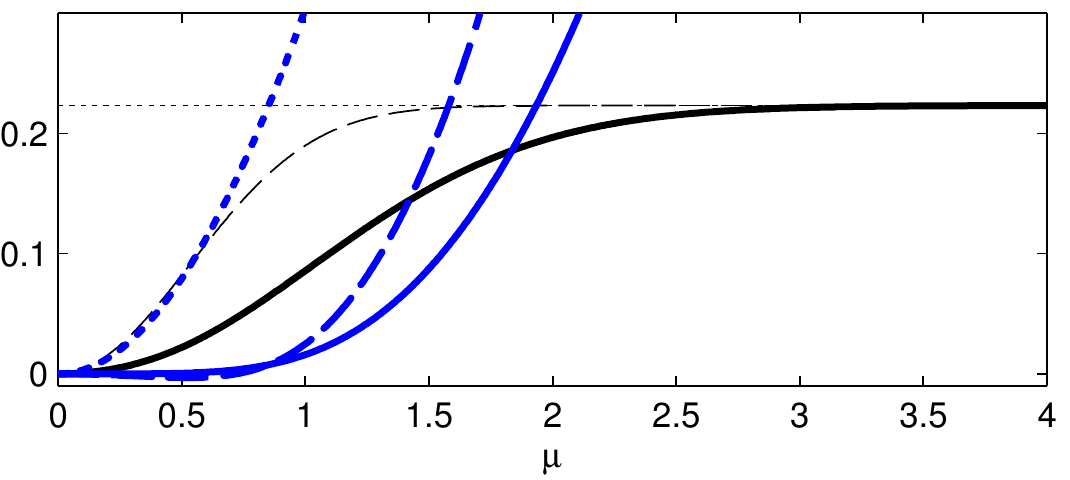}	
		\caption{ The exact divergence $D_{KL}(\hat{p}(\cdot|\mathcal{H}_{12})||p(\cdot))$ (solid blue), the simple approximation given in~\eqref{eq:revAKLtight}  (dotted blue), and the proposed approximation $R(1,2)$ (dashed blue) are plotted for different values of $\mu$.
In addition, the exact pruning divergence $D_{KL}(\hat{p}(\cdot|\mathcal{H}_{02})||p(\cdot))$ (solid black), the crude pruning upper bound $-\log(1-w_2)$ (dotted black) and the proposed pruning upper bound $R(0,2)$ (dashed black) are plotted.}
\label{fig:merginganalysis}
\end{figure}
\end{exmp}

The approximate MRA proposed in this work uses the two approximations  for pruning and merging divergences, $R(I,J)$ where $0\le I\le N$, $1\le J\le N$, and $I \neq J$. We will refer to this algorithm as Approximate Reverse Kullback-Leibler (ARKL) MRA. The precision of ARKL can be traded for less computation time by replacing the approximations for pruning and merging hypotheses by their less precise counterparts given in sections~\ref{sec:pruningapprox} and~\ref{sec:mergingapprox}.

%\begin{align}
%\int{q_{IJ}}&\log\frac{q_{IJ}}{w_Iq_I+w_Jq_J}\d x\nonumber\\
%&\approx \min_\alpha {\left[\int{\alpha q_{IJ}(1-\frac{q_J}{\overline{q_J}})\log\frac{\alpha q_{IJ}}{w_Iq_I}\d x} \right.} \nonumber\\
%&\left.+\int{(1-\alpha)q_{IJ}(1-\frac{q_I}{\overline{q_I}})\log\frac{(1-\alpha)q_{IJ}}{w_Jq_J}\d x}\right]
%\end{align}
%where $0<\alpha<1$ and  $\overline{q_K}\triangleq\max_x q_K(x)$.
%
%\begin{align}
%&\int{\alpha q_{IJ}(1-\frac{q_J}{\overline{q_J}})\log\frac{\alpha q_{IJ}}{w_Iq_I}\d x}  \nonumber\\
%&+\int{(1-\alpha)q_{IJ}(1-\frac{q_I}{\overline{q_I}})\log\frac{(1-\alpha)q_{IJ}}{w_Jq_J}\d x}\nonumber\\
%&=\alpha\log\frac{\alpha}{w_I}\int{ q_{IJ}(1-\frac{q_J}{\overline{q_J}})\d x}  \nonumber\\
%&+\alpha \int{ q_{IJ}(1-\frac{q_J}{\overline{q_J}})\log\frac{q_{IJ}}{q_I}\d x}  \nonumber\\
%&+(1-\alpha)\log\frac{(1-\alpha)}{w_J}\int{q_{IJ}(1-\frac{q_I}{\overline{q_I}})\d x}\nonumber\\
%&+(1-\alpha)\int{q_{IJ}(1-\frac{q_I}{\overline{q_I}})\log\frac{q_{IJ}}{q_J}\d x}\nonumber\\
%\end{align}

\begin{figure*}[!t]%
\centering

\begin{subfigure}{.5\columnwidth}
\includegraphics[width=\columnwidth]{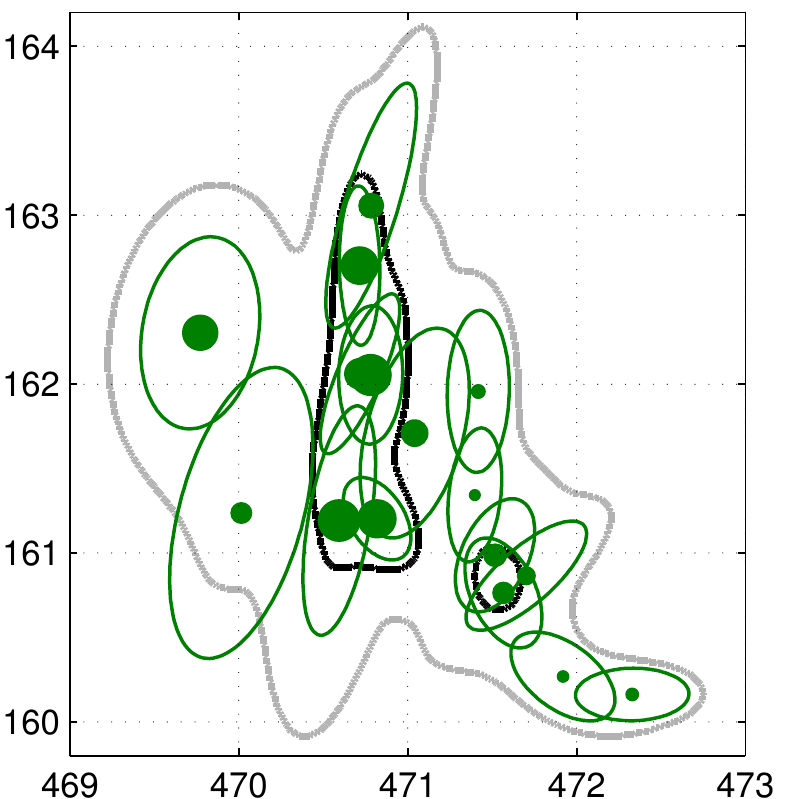}%
\caption{Original density}%
\label{fig:F16o}%
\end{subfigure}%
\begin{subfigure}{.5\columnwidth}
\includegraphics[width=\columnwidth]{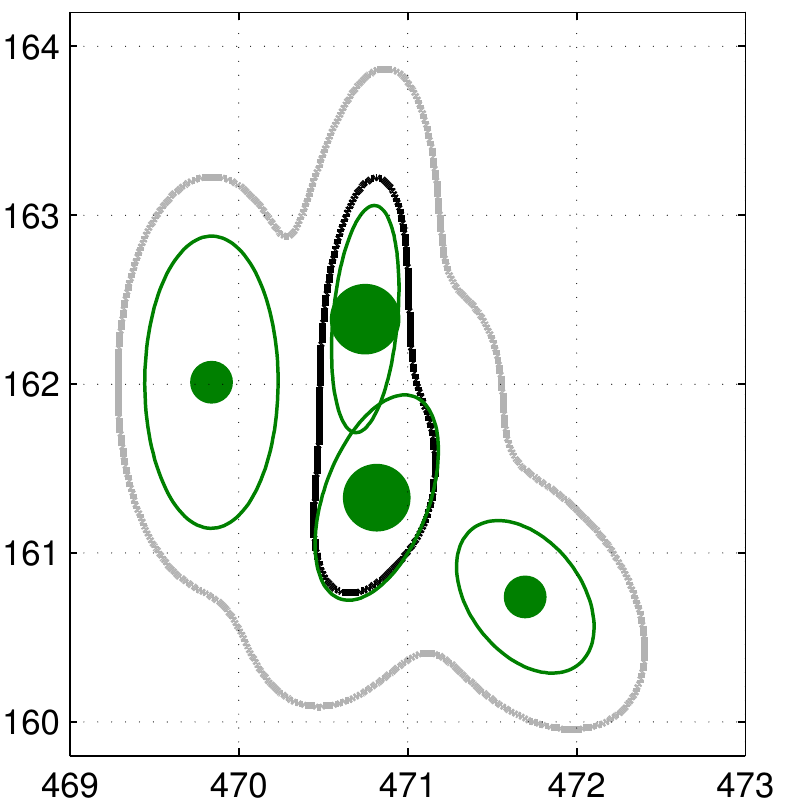}%
\caption{Reduced by Runnalls'}%
\label{fig:akl16}%
\end{subfigure}%
\begin{subfigure}{.5\columnwidth}
\includegraphics[width=\columnwidth]{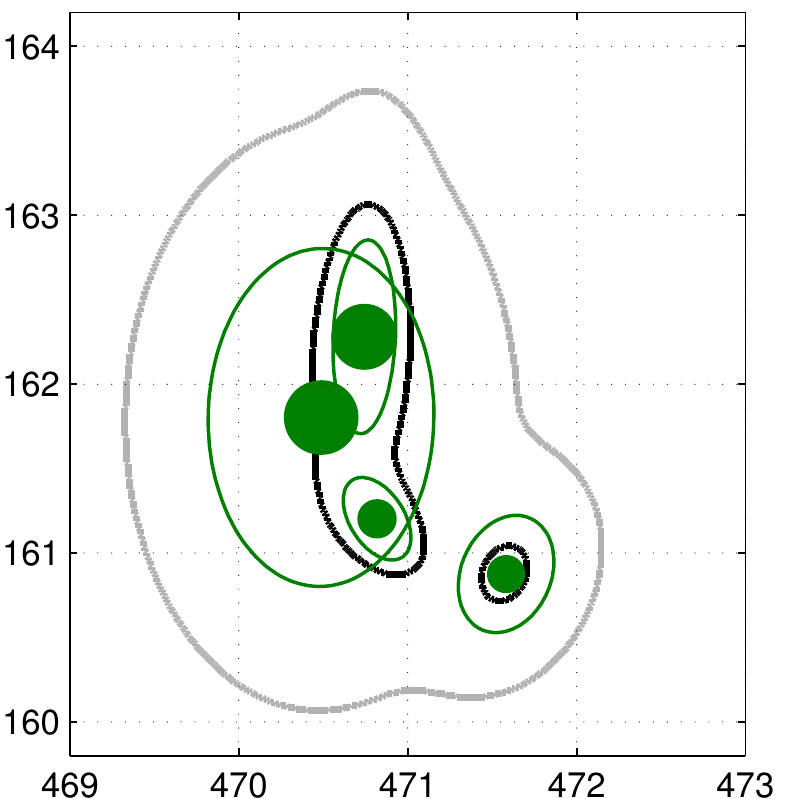}%
\caption{Reduced by Williams'}%
\label{fig:ise16}%
\end{subfigure}%
\begin{subfigure}{.5\columnwidth}
\includegraphics[width=\columnwidth]{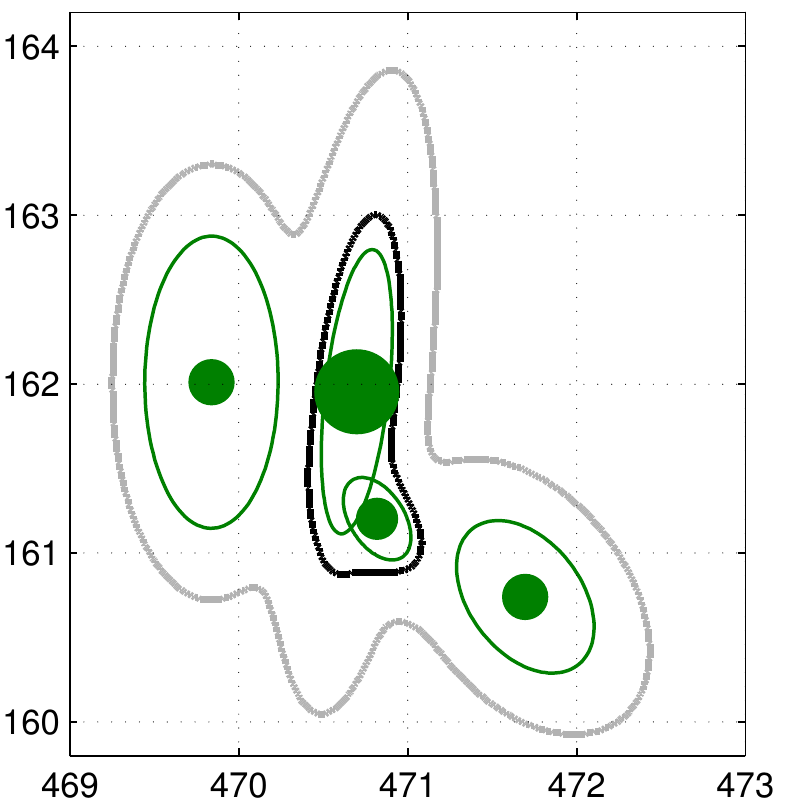}%
\caption{Reduced by ARKL}%
\label{fig:rakl16}%
\end{subfigure}%

\caption{The data for the original mixture density is courtesy of Andrew R. Runnalls and shows the horizontal position errors in the terrain-referenced navigation. The unit of axes is in kilometers and are represented in UK National Grid coordinates. The original mixture in \ref{fig:F16o} is reduced in  \ref{fig:akl16}, \ref{fig:ise16} and \ref{fig:rakl16} using different MRAs.}
\label{fig:F16all}

\end{figure*}
%%=============================================================================
 %--------------------------
\begin{figure*}[!t]%
\centering
\begin{subfigure}{.33\columnwidth}
\includegraphics[width=\columnwidth]{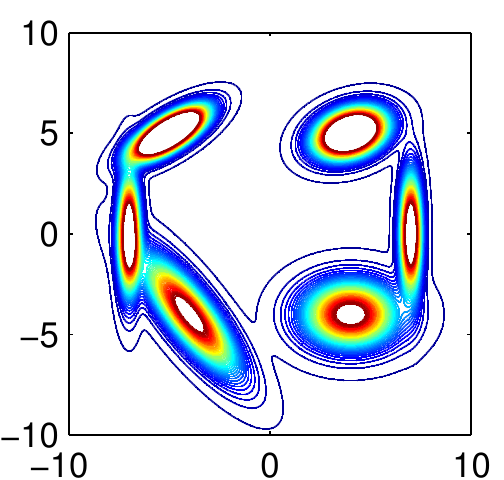}%

\includegraphics[width=\columnwidth]{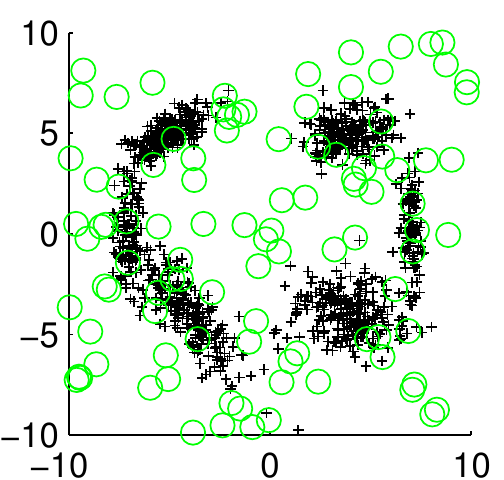}%
\caption{Original density}%
\label{fig:trueclusters}%
\end{subfigure}%
\begin{subfigure}{.33\columnwidth}
\includegraphics[width=\columnwidth]{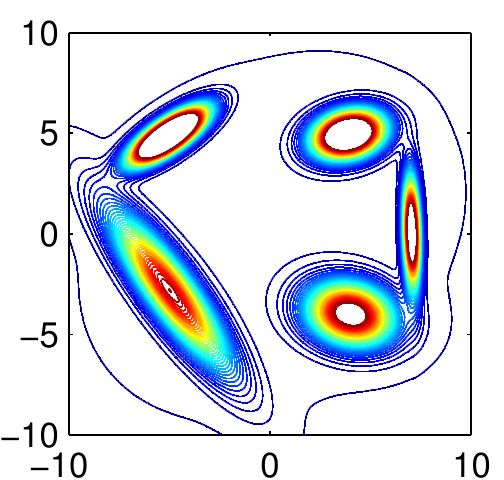}%

\includegraphics[width=\columnwidth]{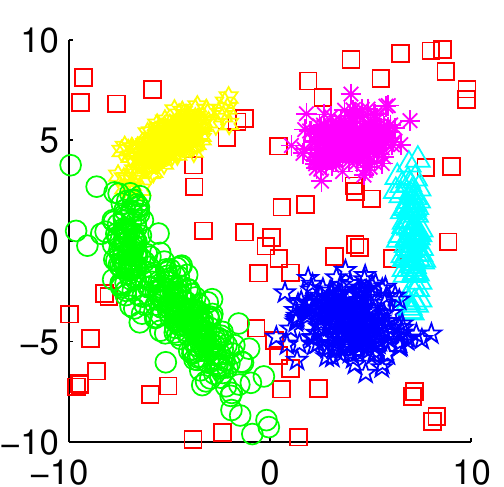}%
\caption{$6$ clusters by EM}%
\label{fig:directEM}%
\end{subfigure}%
\begin{subfigure}{.33\columnwidth}
\includegraphics[width=\columnwidth]{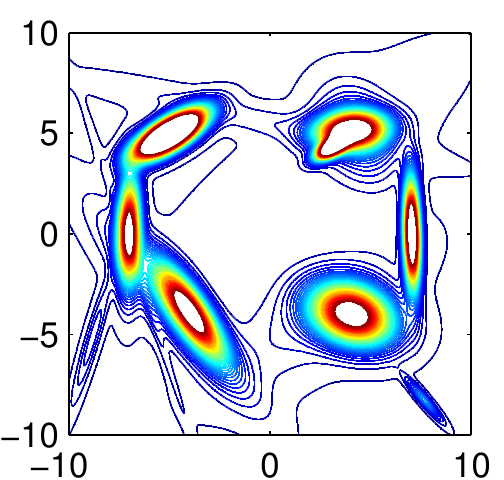}%

\includegraphics[width=\columnwidth]{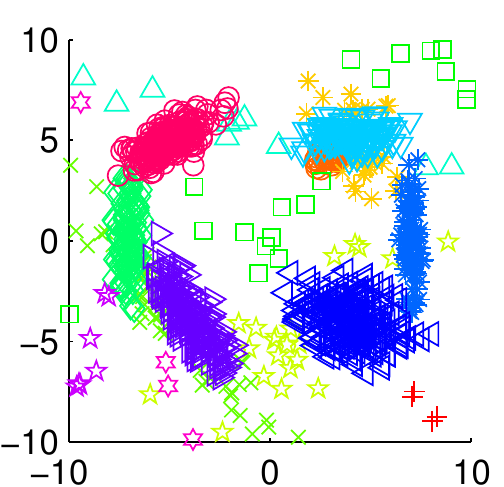}%
\caption{$15$ clusters by EM}%
\label{fig:moreclusters}%
\end{subfigure}%
\begin{subfigure}{.33\columnwidth}
\includegraphics[width=\columnwidth]{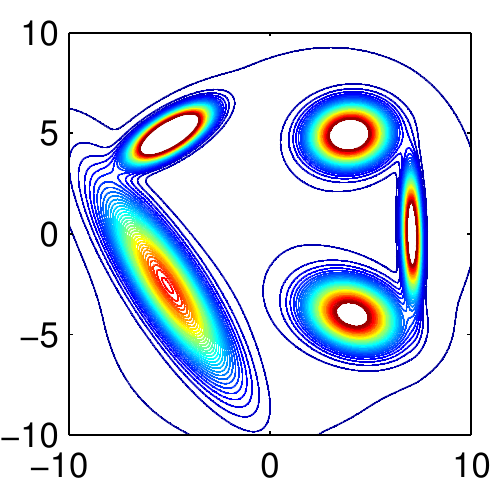}%

\includegraphics[width=\columnwidth]{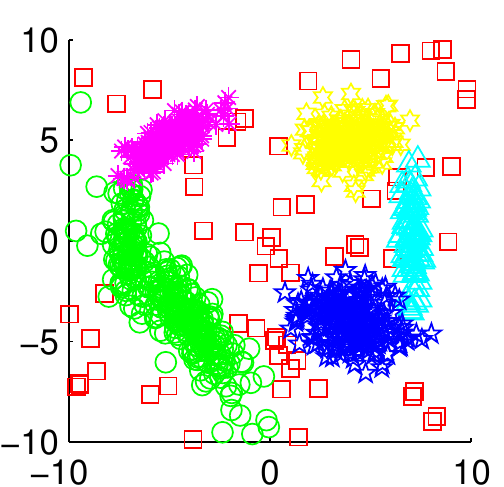}%
\caption{Runnalls'}%
\label{fig:aklclusters}%
\end{subfigure}%
\begin{subfigure}{.33\columnwidth}
\includegraphics[width=\columnwidth]{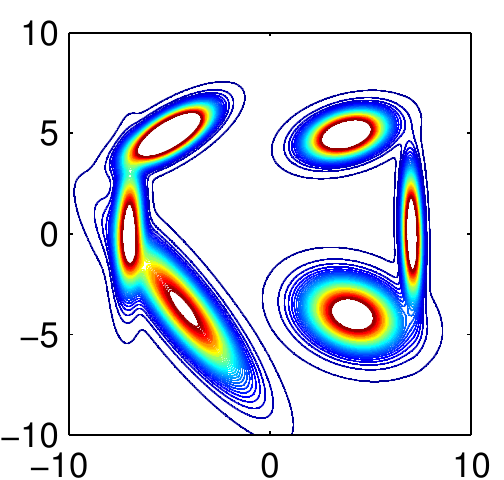}%

\includegraphics[width=\columnwidth]{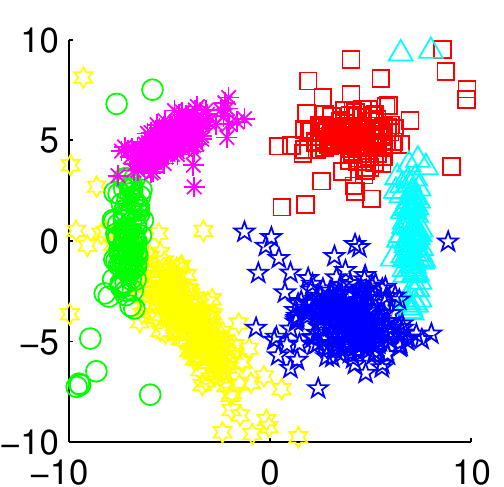}%
\caption{ Williams'}%
\label{fig:iseclusters}%
\end{subfigure}%
\begin{subfigure}{.33\columnwidth}
\includegraphics[width=\columnwidth]{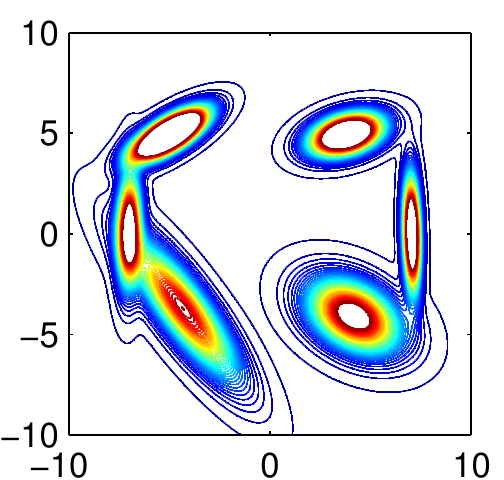}%

\includegraphics[width=\columnwidth]{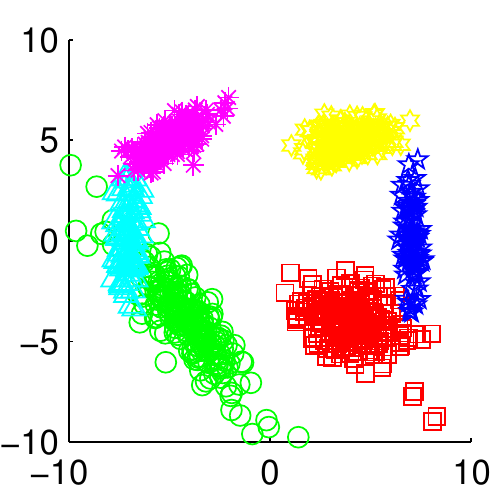}%
\caption{ ARKL}%
\label{fig:raklclusters}%
\end{subfigure}%
\caption{Results of the clustering example. In the top row, the contour plots of the GMs are given. The bottom row presents the corresponding data points associated to the clusters in different colors.}
\label{fig:clusters}
\end{figure*}
 %--------------------------
%\begin{figure*}%
%\centering
%\begin{subfigure}{.33\columnwidth}
%\includegraphics[width=\columnwidth]{true_clusters_data.pdf}%
%\caption{Original density}%
%\label{fig:trueclusters}%
%\end{subfigure}%
%\begin{subfigure}{.33\columnwidth}
%\includegraphics[width=\columnwidth]{directEM_data.pdf}%
%\caption{2 clusters by EM}%
%\label{fig:directEM}%
%\end{subfigure}%
%\begin{subfigure}{.33\columnwidth}
%\includegraphics[width=\columnwidth]{moreclusters_data.pdf}%
%\caption{7 clusters by EM}%
%\label{fig:moreclusters}%
%\end{subfigure}%
%\begin{subfigure}{.33\columnwidth}
%\includegraphics[width=\columnwidth]{aklclusters_data.pdf}%
%\caption{Reduced by AKL}%
%\label{fig:aklclusters}%
%\end{subfigure}%
%\begin{subfigure}{.33\columnwidth}
%\includegraphics[width=\columnwidth]{iseclusters_data.pdf}%
%\caption{Reduced by ISE}%
%\label{fig:iseclusters}%
%\end{subfigure}%
%\begin{subfigure}{.33\columnwidth}
%\includegraphics[width=\columnwidth]{raklclusters_data.pdf}%
%\caption{Reduced by ARKL}%
%\label{fig:raklclusters}%
%\end{subfigure}%
%\caption{ A clustering example using EM.  }
%\label{fig:clusters}
%\end{figure*}

%%=============================================================================
%%===============================  New SECTION  ===============================
%%=============================================================================
\section{Simulation Results}
\label{sec:numsim}
In this section, the performance of the proposed MRA, ARKL, is compared with Runnalls' and Williams' algorithms in two examples. In the first example, we use the real world data from  terrain-referenced navigation. In the second example, we evaluate the MRAs on a clustering problem with simulated data.
%%=============================================================================
\subsection{Example with real world data}
\label{sec:F16}
An example that Runnalls used in \cite[Sec.VIII]{KLrunnalls2007} is repeated here for the comparison of ARKL with Runnalls' and Williams' algorithms. In this example a Gaussian Mixture (GM) consisting of $16$ components which represents a terrain-referenced navigation system's state estimate is reduced to a GM with $4$ components. The random variable whose distribution is represented by a GM is $15$ dimensional  but only two elements of it (horizontal components of position error) are visualized in \cite[Sec.VIII]{KLrunnalls2007}. We use the marginal distribution of the  visualized random variables to illustrate the differences of MRAs.

Figure~\ref{fig:F16o} shows the original mixture density. Each Gaussian component is represented by its mean, its weight, which is proportional to the area of the marker representing the mean, and its $50$ percentile contour (an elliptical contour enclosing the $50\%$ probability mass of the Gaussian density).  The thick gray curves in Figure~\ref{fig:F16all} are the $95$ percentile contour and the thick black curves show the $50$ percentile contour of the mixture density.

In Figures~\ref{fig:akl16} and~\ref{fig:ise16}, the reduced GMs using Runnalls' and Williams' methods are illustrated by their components and their $50$ and $95$ percentile contours\footnote{The difference between Figures~\ref{fig:akl16} and~\ref{fig:ise16} and~\cite[Fig.2]{KLrunnalls2007} is because of the exclusion of the pruning hypotheses in implementation of Williams' algorithm in \cite[Fig.2]{KLrunnalls2007} and the fact that we are only using the marginal distribution of the two position error elements in the random variable.}.  Figure~\ref{fig:rakl16} illustrates the result of the ARKL algorithm.

Williams' algorithm is the only algorithm  that preserves the second mode in the $50$ percentile contour, but has distorted the $95$ percentile contour the most. The $50$ percentile contour of ARKL is the most similar to the original density. Also, the shape of the $95$ percentile contour is well preserved by  ARKL as well as Runnalls'. An interesting observation is that the $95$ percentile contour for ARKL has sharper corners than that obtained by  Runnalls' algorithm which is a manifestation of the pruning decisions (zero-forcing) in ARKL as opposed to the merging (zero-avoiding) decisions in Runnalls' algorithm for small-weight components.

%Perhaps, the proportionality of the two contours for two MRAs where KLD is used as the divergence is due to the definition of the KLD. The KLD in both directions concerns with the proportion of the probability densities while ISE is concerned with the absolute difference of the two probability densities and this absolute difference is uniformly penalized regardless of the value of the argument densities of the ISE.

The MRAs are compared quantitatively in terms of divergences, FKLD, RKLD and ISE in Table~\ref{table:F16}. The FKLD and RKLD are calculated using MC integration while ISE is calculated using its analytical expression. Naturally, MRAs obtain the least divergence with their respective criteria. That is, Runnalls' algorithm obtains the lowest FKLD, Williams' algorithm obtains the lowest ISE, and ARKL obtains the lowest RKLD. In terms of ISE, Runnalls' algorithm and ARKL seems to be very close to each other which implies that for large weight components (large probability mass), these algorithms work similarly. In terms of FKLD, the Williams' algorithm obtains much closer a value to that of Runnalls' algorithm than ARKL which means that the merging (zero-avoiding) properties of Williams' algorithm is closer to Runnalls' algorithm. Finally, in terms of RKLD, Williams' algorithm and Runnalls' algorithm have similar and larger values than ARKL. This makes us conclude that pruning (zero-forcing) ability of Williams' algorithm is as bad as Runnalls' algorithm for this example.
%--------------------------
\begin{table}[t]
\caption{Comparison of MRAs in terms of the divergence between the reduced mixture density and the original density.}
\centering
\begin{tabular}{l ccc }
\toprule
  &\multicolumn{3}{c} {\textbf{Mixture Reduction Algorithms}}    \\
\cmidrule(lr){2-4}
\textbf{Divergence Measures} &\textbf{Williams'}&\textbf{Runnalls'}&\textbf{ARKL}\\\midrule
\textbf{ISE}&0.0175&0.0255&0.0271\\
\textbf{FKLD}&0.1177&0.0665&0.3024\\
\textbf{RKLD}&0.1325&0.1234&0.0340\\
\bottomrule
\end{tabular}
\label{table:F16}
\end{table}
%--------------------------

%%=============================================================================

%%=============================================================================
\subsection{Robust clustering} \label{sec:robustclustering}
In the following, we illustrate the necessity of the pruning capability of an MRA (as well as merging) using a clustering example in an environment with spurious measurements.

Consider a clustering problem where the number of clusters $N_c$ is known a priori. Also, assume that the data is corrupted by a small number of spurious measurements, which are the data points not belonging to any of the true clusters and can be considered to be outliers. The clustering task is to retrieve $N_c$ clusters from the corrupted data.

A straightforward approach to this problem is to neglect the existence of the spurious measurements and cluster the corrupted data directly into $N_c$ clusters. Since this solution does not take the spurious measurements into account, the resulting clusters might not represent the true data satisfactorily. Another solution to the problem is to first cluster the data points into $N>N_c$ clusters and then reduce the number of clusters to $N_c$. When GMs are used for clustering, an appropriate MRA can be used to reduce the number of clusters (component densities) from $N$ to $N_c$. An MRA which can prune components with small weights which are distant from the rest of the probability mass is a good candidate. On the other hand, an MRA which reduces the mixture using only merging operations is not preferable because the spurious measurements would corrupt the true clusters. Furthermore, when a component is being pruned/discarded the data points associated to the pruned component density can also be discarded as potential spurious data points. Consequently, it is expected that ARKL and Williams' algorithm might perform better than Runnalls' algorithm in this example because of their pruning capability.

In this simulation $n=1000$ data points are generated from a Gaussian mixture with $N_c=6$ components, $\{x_i\}_{i=1}^n\thicksim p(x)=\sum_{j=1}^{N_c} w_j\N(x;\mu_j,\Sigma_j)$ where the parameters of $p(x)$ are given in Table~\ref{table:MixParams}.
%\begin{align*}
%w&=[w_1,\cdots,w_{N_c}]=\left[\frac{1}{5},\frac{1}{5},\frac{1}{5},\frac{1}{5},\frac{1}{10},\frac{1}{10}\right]\\
%\mu&=[\mu_1,\cdots,\mu_{N_c}]=\left[ \begin{matrix}
 %-5 &   4  &   4  & -4  & -7  &  7 \\
 %5  &  5   & -4  & -4  &  0  &  0 \end{matrix}\right]\\
%\Sigma&=[\Sigma_1,\cdots,\Sigma_{N_c}]\\
%&=\begin{bmatrix}
%1           &  \frac{1}{2}  &    1  &   \frac{1}{5} &    2 &   0  &  2 &  -2  & \frac{1}{10} & 0 & \frac{1}{10} & 0\\
%\frac{1}{2} &  \frac{1}{2}  &   \frac{1}{5} &  \frac{1}{2}    &    0  &  1  & -2     &    3   &      0 & 3  &    0 &   3
%\end{bmatrix}
%\end{align*}
%=====================
\begin{table}[ht]
\centering
\caption{The  parameters of the original mixture density used in the example of  section~\ref{sec:robustclustering} and presented in Figure~\ref{fig:trueclusters}. }
\label{table:MixParams}
\begin{tabular}{cccc  cccc}
\toprule
\multicolumn{4}{c} {$1\leq j \leq 3$} &\multicolumn{4}{c} {$4\leq j \leq 6$}   \\
\cmidrule(lr){1-4} \cmidrule(lr){5-8}

$j$& $w_j$ & $\mu_j$ & $\Sigma_j$&$j$& $w_j$ & $\mu_j$ & $\Sigma_j$\\
\cmidrule(lr){1-4} \cmidrule(lr){5-8}

$1$ & $0.2$ &  $\begin{bmatrix} -5 \\ 5\end{bmatrix} $& $\begin{bmatrix}1  &  0.5\\[3pt] 0.5 &  0.5\end{bmatrix}$ &$4$ & $0.2$ &  $\begin{bmatrix} -4 \\[3pt] -4\end{bmatrix} $& $\begin{bmatrix}2  &  -2\\[3pt] -2 & 3\end{bmatrix}$ \\

\rule{0pt}{4ex}
$2$ & $0.2$ &  $\begin{bmatrix} 4 \\5\end{bmatrix} $& $\begin{bmatrix}1  &  0.2\\[3pt] 0.2 &  0.5\end{bmatrix}$ &$5$ & $0.1$ &  $\begin{bmatrix} -7 \\0\end{bmatrix} $&$\begin{bmatrix}0.1  &  0\\ 0 &  3\end{bmatrix}$\\

\rule{0pt}{4ex}
$3$ & $0.2$ &  $\begin{bmatrix} 4 \\-4\end{bmatrix} $& $\begin{bmatrix}2  &  0\\ 0 &  1\end{bmatrix}$ &$6$ & $0.1$ &  $\begin{bmatrix} 7 \\ 0\end{bmatrix} $& $\begin{bmatrix}0.1  &  0\\[3pt] 0 &  3\end{bmatrix}$ \\
\bottomrule
\end{tabular}
\end{table}
%====================
The data points are augmented with $m=100$ spurious data points sampled uniformly from a multivariate uniform distribution over the square region $\mathcal{A}$, i.e., $\{y_i\}_{i=1}^m\thicksim \mathcal{U}(\mathcal{A})$, where the center of the square is at the origin and each side is $20$ units long. The clustering results of the following methods are presented.
 \begin{itemize}
 \item The direct clustering of the corrupted data, i.e., $z=\{\{x_i\}_{i=1}^n,\{y_i\}_{i=1}^m\}$  into 6 clusters using  Expectation Maximization (EM) algorithm~\cite[Chapter~9]{bishop2006}.
 \item The clustering of the corrupted data into 15 clusters using  EM algorithm.
 \item The clustering of the corrupted data into 15 clusters using EM algorithm and then reduction to 6 clusters using
 \begin{itemize}
 \item Runnalls' algorithm
 \item Williams' algorithm
 \item ARKL
 \end{itemize}
\end{itemize}

In Figure~\ref{fig:clusters} the contour plots of the densities along with the corresponding data points are illustrated. In Figure~\ref{fig:trueclusters} (top) the contour plots of the true clusters are plotted. The true (black) and spurious (green) data points are shown in Figure~\ref{fig:trueclusters} (bottom).

In Figure~\ref{fig:directEM}, the contour plot (top) of the GM obtained by EM algorithm clustering the corrupted data directly into $6$ clusters is shown with the corresponding cluster data points (bottom). The direct clustering of the corrupted data into $6$ components results in a clustering which does not capture all of the true clusters since a spurious cluster with a large covariance is formed out of spurious data points (red data points).

As a remedy, the data is first clustered into $15$ clusters (see Figure~\ref{fig:moreclusters}) using EM algorithm and the resulting 15-component GM is then reduced to $6$ components using Runnalls' algorithm (Figure~\ref{fig:aklclusters}), Williams' algorithm (Figure~\ref{fig:iseclusters}) and ARKL (Figure~\ref{fig:raklclusters}). With these algorithms, when a component is pruned (which might be the case for Williams' algorithm and ARKL), the data points associated with that component are discarded as well. On the other hand, when a component is merged with another component, data points associated with the merged component are reassigned to the updated set of clusters according to their proximity measured by Mahalanobis distance.

The final clustering obtained by Runnalls' algorithm is almost the same as the result obtained by the EM algorithm with 6 clusters. This observation is in line with the Maximum Likelihood interpretation/justification of the Runnalls' algorithm given in~\cite{KLrunnalls2007}. Williams' algorithm and ARKL capture the shape of the original GM and the $6$ clusters much better than Runnalls' algorithm because these two algorithms can prune (as well as merge) while Runnalls' algorithm can only merge. While some of the spurious clusters/measurements are merged with the true clusters by Williams' algorithm, almost all spurious clusters/measurements are pruned by ARKL.
Lastly, as aforementioned, the computation complexity of Williams' method is $\mathcal{O}(N^4)$ while reducing a mixture with $N$ components to a mixture with $N-1$ components. However, Runnalls' and the ARKL methods have $\mathcal{O}(N^2)$ complexity, which makes them preferable when a fast reduction algorithm is needed for reducing a mixture with many components.

%\subsection{Target Tracking example}
%\label{sec:1DPHD}
%

%%=============================================================================
%%===============================  New SECTION  ===============================
%%=============================================================================

\section{Conclusion}
\label{sec:conclusion}
We investigated using the reverse Kullback-Leibler divergence as the cost function for greedy mixture reduction. Proposed analytical approximations for the cost function results in an efficient algorithm which is capable of both merging and pruning components in a mixture.  This property can help preserving the peaks of the original mixture after the reduction and it is missing in FKLD based methods. We compared the method with two well-known mixture reduction algorithms and illustrated its advantages both in simulated and real data scenarios.

%An algorithm for reducing Gaussian mixture densities is proposed which chooses the best reduction hypothesis among the merging and pruning hypotheses via minimizing a cost which approximates the reverse Kullback-Leibler divergence. It is also shown that an algorithm that minimizes the  reverse Kullback-Leibler divergence can select among the pruning hypotheses as well as the merging hypotheses. Not only is   the  proposed algorithm motivated and compared to two other algorithms  via analytical examples and numerical simulations, but also is easy to motivate on an intuitive level; It merges two components when they are close and prunes the components that are far from the rest of the probability mass and have a relatively small weight.
%%=============================================================================
%%===============================  New SECTION  ===============================
%%=============================================================================

\bibliographystyle{IEEEtran}
\bibliography{IEEEabrv,RKLD_arxiv}
%%=============================================================================
%%===============================  New SECTION  ===============================
%%=============================================================================
\appendix
%%=============================================================================
\subsection{Proof of Lemma~\ref{lem:vub}}
\label{sec:proofoflem:vub}
In the proof of Lemma~\ref{lem:vub} we use the same approach as the one used in \cite[Sec. 8]{hershey2007}.
Let  $0<\alpha<1$. Using the log-sum inequality we can obtain an upper bound on the left hand side of~\eqref{eq:basicintegral} as
\begin{align}
& \int{q_K\log\frac{q_K}{w_Iq_I+w_Jq_J}\d x} \nonumber\\
&=\int{({\alpha q_K+(1-\alpha) q_K})\log\frac{{\alpha q_K+(1-\alpha) q_K}}{w_Iq_I+w_Jq_J}\d x}\nonumber\\
&\leq \int{\alpha q_K\log\frac{{\alpha q_K}}{w_Iq_I}\d x}+\int{(1-\alpha) q_K\log\frac{(1-\alpha) q_K}{w_Jq_J}\d x}\nonumber\\
&=\alpha\int{ q_K\log\frac{{ q_K}}{q_I}\d x}+\alpha\log\frac{{\alpha }}{w_I}\nonumber\\
&+  (1-\alpha)\int{ q_K\log\frac{{ q_K}}{q_J}\d x} +   (1-\alpha) \log\frac{1-\alpha}{w_J}\nonumber\\
&= \alpha D_{KL}(q_K||q_I) +\alpha\log\frac{{\alpha }}{w_I}+ (1-\alpha) D_{KL}(q_K||q_J)     \nonumber\\
&+ (1-\alpha) \log\frac{1-\alpha}{w_J}. \label{eq:lemvubcost}
\end{align}
The upper bound can be minimized with respect to $\alpha$ to obtain the best upper bound.
The minimum is obtained by taking the derivative of \eqref{eq:lemvubcost} with respect to $\alpha$; equating the result to zero and solving for $\alpha$, which gives
\begin{equation*}
\alpha^*=\frac{w_I\exp(-D_{KL}(q_K||q_I))}{w_I\exp(-D_{KL}(q_K||q_I))+w_J\exp(-D_{KL}(q_K||q_J))}.
\end{equation*}
Substituting $\alpha^*$ into ~\eqref{eq:lemvubcost} gives the upper bound in~\eqref{eq:basicintegral}.
%%=============================================================================
%\subsection{Proof of Lemma~\ref{lem:vfup}}
%\label{sec:proofoflem:vfup}
%Consider the following function,
%\begin{equation}
%\label{eq:Ffunctional}
%F(\alpha)=\alpha\log(\phi\alpha)+(1-\alpha)\log(\psi(1-\alpha))
%\end{equation}
%where $\phi$ and $\psi$ are positive constants. By taking the derivative of $F(\cdot)$ with respect to $\alpha$ and equating the result by zero and solving for $\alpha$ we obtain
%\begin{equation}
%\label{eq:functionalsolution}
%\widehat{\alpha}=\frac{1}{1+\frac{\phi}{\psi}}.
%\end{equation}
%By inserting the functions $\phi(x)=\frac{q_{K}(x)}{w_Iq_I(x)}$ and $\psi(x)=\frac{q_{K}(x)}{w_Jq_J(x)}$ in \eqref{eq:Ffunctional} and \eqref{eq:functionalsolution} the proof follows.
%%=============================================================================
\subsection{Derivation of $V(q_K,q_I,q_J)$}
\label{sec:derivationofV}
Let $q_K(x)=\N(x;\mu_K,\Sigma_K)$, $q_I(x)=\N(x;\mu_I,\Sigma_I)$ and $q_J(x)=\N(x;\mu_J,\Sigma_J)$ and $x\in\mathbb{R}^k$. The integral $V(q_K,q_I,q_J)$ can be written as
\begin{align*}
V&(q_K,q_I,q_J)\triangleq\int{q_{K}\left(1-\frac{q_I}{q_I^{max}}\right)\log\frac{q_{K}}{q_J}\d x}\\
=&D_{KL}(q_{K}||q_J)-\frac{1}{q_I^{max}}\int{q_{K}{q_I}\log\frac{q_{K}}{q_J}\d x}\\
=&D_{KL}(q_{K}||q_J)-(2\pi)^{k/2}|\Sigma_I|^{1/2}\N(\mu_I;\mu_{K},\Sigma_{K}+\Sigma_I)\times\\
&\left(\int{\N(x;\mu^*,\Sigma^*)\log{q_{K}}\d x}-\int{\N(x;\mu^*,\Sigma^*)\log{q_J}\d x}\right)\\
=&0.5\log\frac{|\Sigma_K|}{|\Sigma_J|} \\
&-0.5\Tr{[\Sigma_J^{-1}(\Sigma_J-\Sigma_K-(\mu_J-\mu_K)(\mu_J-\mu_K)^\t)]}\\
&-(2\pi)^{k/2}|\Sigma_I|^{1/2}\N(\mu_I;\mu_{K},\Sigma_{K}+\Sigma_I)\big[  -0.5\log{|2\pi\Sigma_K|}\\
&-0.5\Tr{[\Sigma_K^{-1}(\Sigma^*+(\mu_K-\mu^*)(\mu_K-\mu^*)^\t)]} \\
&+0.5\log{|2\pi\Sigma_J|}\nonumber\\
& +0.5\Tr{[\Sigma_J^{-1}(\Sigma^*+(\mu_J-\mu^*)(\mu_J-\mu^*)^\t)]}\big].
\end{align*}
where we use the following well-known results for Gaussian distributions.
\begin{itemize}
\item Divergence
\begin{align}
&D_{KL}(\N(x;\mu_K,\Sigma_K)||\N(x;\mu_J,\Sigma_J))=0.5\log\frac{|\Sigma_K|}{|\Sigma_J|} \nonumber\\
&-0.5\Tr{[\Sigma_J^{-1}(\Sigma_J-\Sigma_K-(\mu_J-\mu_K)(\mu_J-\mu_K)^\t)]}.
\end{align}
\item Multiplication
\begin{align}
&\N(x;\mu_I,\Sigma_I)\N(x;\mu_K,\Sigma_K)=\N(\mu_K;\mu_I,\Sigma_I+\Sigma_K)\nonumber\\
&\hspace{4cm}\times\N(x;\mu^*,\Sigma^*)
\end{align}
where
\begin{align}
\mu^*&=\mu_I+\Sigma_I(\Sigma_I+\Sigma_K)^{-1}(\mu_K-\mu_I),\\
\Sigma^*&=\Sigma_I-\Sigma_I(\Sigma_I+\Sigma_K)^{-1}\Sigma_I.
\end{align}
\item Expected Logarithm
\begin{align}
\int& \N(x;\mu_I,\Sigma_I)\log \N(x;\mu_J,\Sigma_J)\d x=-0.5\log{|2\pi\Sigma_J|} \nonumber\\
&-0.5\Tr{[\Sigma_J^{-1}(\Sigma_I+(\mu_J-\mu_I)(\mu_J-\mu_I)^\t)]}.
\end{align}
\item Maximum Value
\begin{align}
&q_I^{max}=(2\pi)^{-k/2}|\Sigma_I|^{-1/2}.
\end{align}
\end{itemize}

\vfill

\end{document}